\documentclass{article} 
\usepackage{iclr2024_conference,times}

\usepackage{amsmath,amsfonts,bm}

\def\eqref#1{equation~\ref{#1}}

\def\1{\bm{1}}

\DeclareMathAlphabet{\mathsfit}{\encodingdefault}{\sfdefault}{m}{sl}
\SetMathAlphabet{\mathsfit}{bold}{\encodingdefault}{\sfdefault}{bx}{n}

\usepackage{tikz}
\usepackage{hyperref}
\usepackage{url}
\usepackage{colortbl}
\usepackage{hyperref}
\usepackage[OT1]{fontenc}
\usepackage{pifont}
\usepackage{graphicx} 
\usepackage{subcaption}
\usepackage{wrapfig}
\usepackage{multirow}
\usepackage{booktabs}
\usepackage{xcolor}
\usepackage{color}       
\usepackage[misc]{ifsym}
\usepackage{amssymb}
\definecolor{best}{rgb}{1.0, 0.6, 0}
\definecolor{best2}{rgb}{1.0, 0.8, 0.6}

\newcommand{\customfootnotetext}[1]{%
  \begingroup
    \renewcommand{\thefootnote}{}
    \footnotetext{#1}%
  \endgroup
}

\title{\textbf{Lotus}: Diffusion-based Visual Foundation Model for High-quality Dense Prediction}

\author{\textbf{Jing He\textsuperscript{1\textcolor{red}{\ding{81}}}}
\textbf{Haodong Li\textsuperscript{1\textcolor{red}{\ding{81}}}}
\textbf{Wei Yin\textsuperscript{2}}
\textbf{Yixun Liang\textsuperscript{1}}
\textbf{Leheng Li\textsuperscript{1}}
\textbf{Kaiqiang Zhou\textsuperscript{3}}
\textbf{Hongbo Zhang\textsuperscript{3}}\\
\textbf{Bingbing Liu\textsuperscript{3}}
\textbf{Yingcong Chen\textsuperscript{1,4 \Letter}}\\
$^{1}$HKUST(GZ) $^{2}$University of Adelaide
$^{3}$Noah's Ark Lab $^{4}$HKUST\\
{
\fontfamily{cmtt}\selectfont\{jhe812, hli736\}@connect.hkust-gz.edu.cn;  yingcongchen@ust.hk
} 
}

\newcommand{\haodong}[1]{#1}
\newcommand{\hd}[1]{#1}
\iclrfinalcopy 
\begin{document}

\maketitle
\vspace{-10mm}

\begin{figure}[h]
    \centering
    \includegraphics[width = 0.99\linewidth]{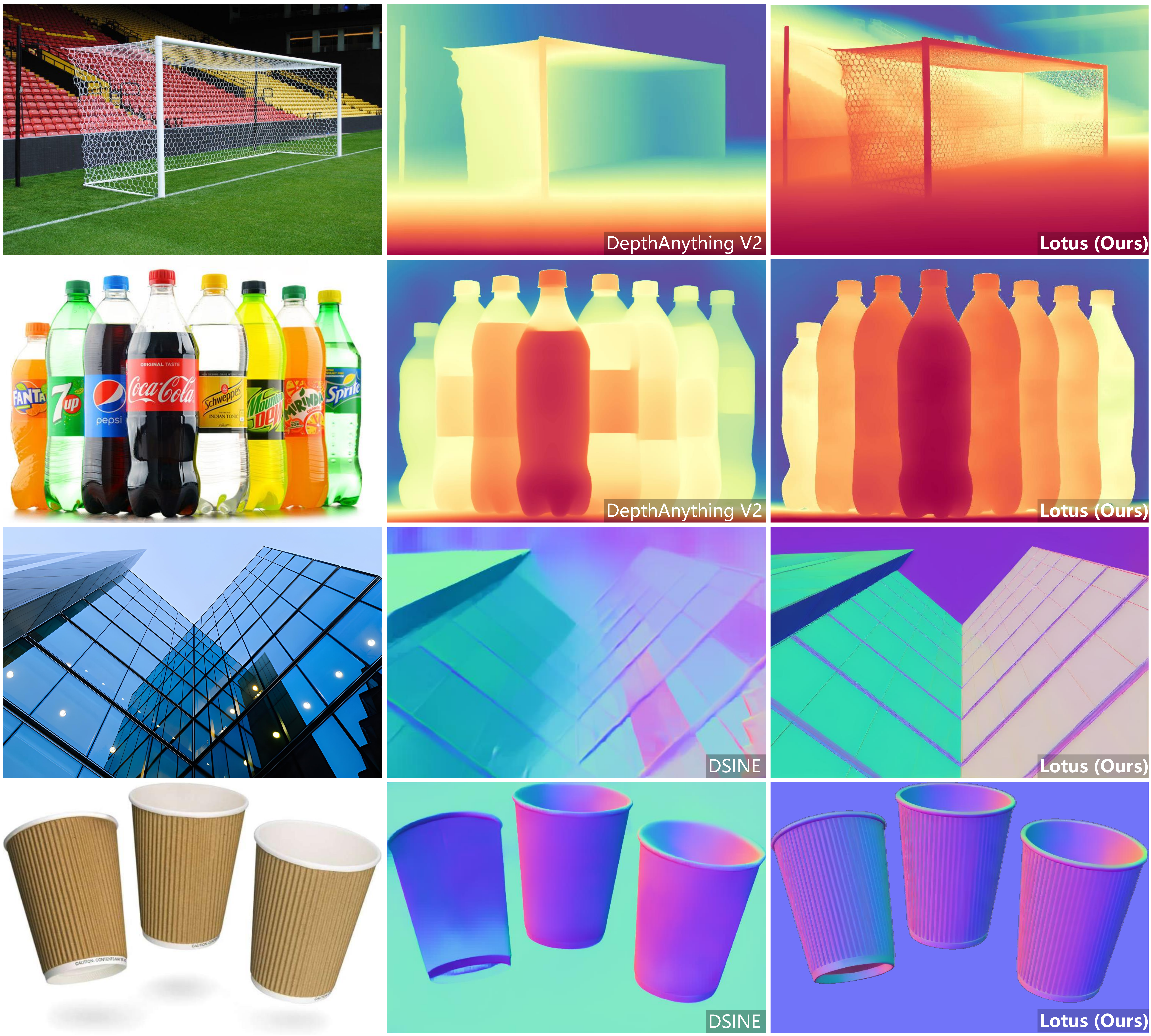}
    \includegraphics[width = 0.95\linewidth]{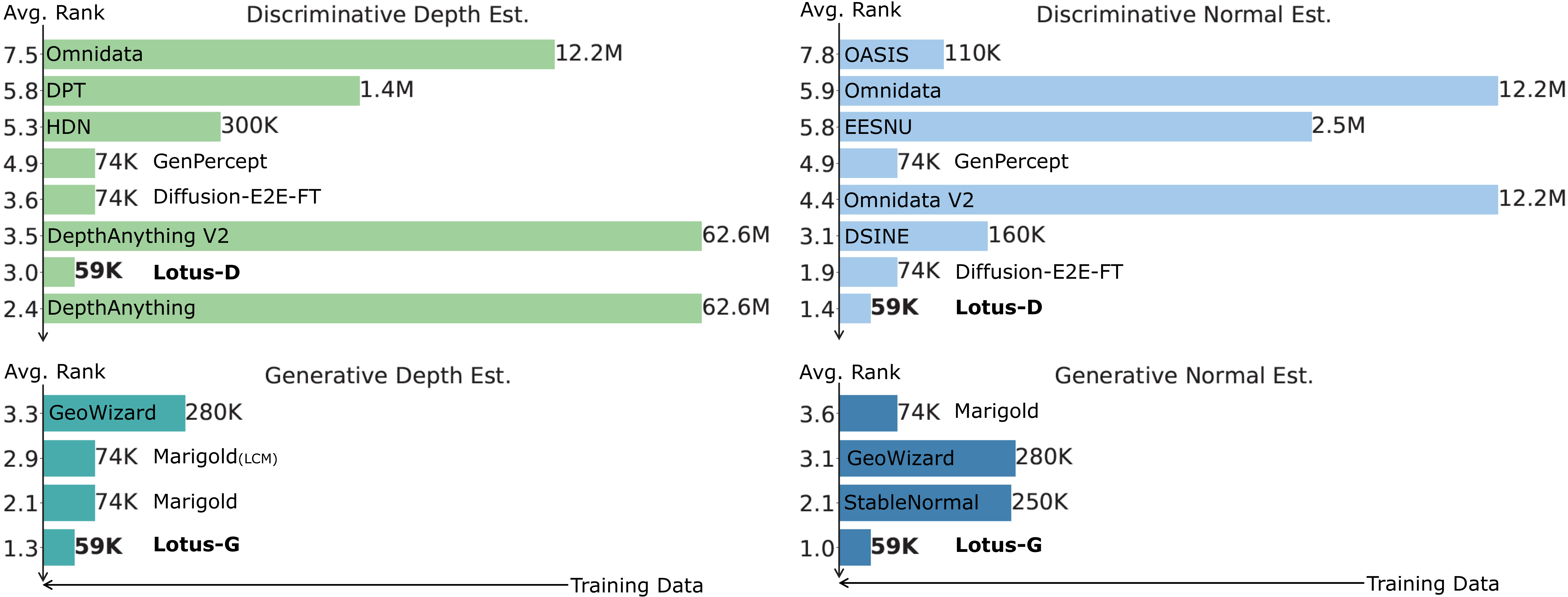}
    \caption{
    We present \textbf{Lotus}, a diffusion-based visual foundation model for dense geometry prediction. With minimal training data, Lotus achieves promising performance in zero-shot depth and normal estimation.
    ``Avg. Rank'' indicates the average ranking across all metrics, where lower values are better. Bar length represents the amount of training data used.
    }
    \label{fig:teaser}
\end{figure}

\customfootnotetext{\textsuperscript{\textcolor{red}{\ding{81}}}Both authors contributed equally (order randomized). \textsuperscript{\Letter} Corresponding author.}

\begin{abstract}

Leveraging the visual priors of pre-trained text-to-image diffusion models offers a promising solution to enhance zero-shot generalization in dense prediction tasks.
\haodong{However, existing methods often uncritically use the original diffusion formulation, which may not be optimal due to the fundamental differences between dense prediction and image generation.}
In this paper, we provide a systemic analysis of the diffusion formulation for the dense prediction, focusing on both quality and efficiency. And we find that the original parameterization type for image generation, which learns to predict noise, is harmful for dense prediction; the multi-step noising/denoising diffusion process is also unnecessary and challenging to optimize.
Based on these insights, we introduce \textbf{Lotus}, a diffusion-based visual foundation model with a simple yet effective adaptation protocol for dense prediction.
\haodong{Specifically, Lotus is trained to directly predict annotations instead of noise, thereby avoiding harmful variance. We also reformulate the diffusion process into a single-step procedure, simplifying optimization and significantly boosting inference speed.
Additionally, we introduce a novel tuning strategy called detail preserver, which achieves more accurate and fine-grained predictions.}
\haodong{Without scaling up the training data or model capacity, Lotus achieves promising performance in zero-shot depth and normal estimation across various datasets. 
It also enhances efficiency, being significantly faster than most existing diffusion-based methods.}
\haodong{Lotus' superior quality and efficiency also enable a wide range of practical applications, such as joint estimation, single/multi-view 3D reconstruction, etc.}
Project page: \href{https://lotus3d.github.io/}{\textcolor{blue}{\fontfamily{cmtt}\selectfont{lotus3d.github.io}}}.
\end{abstract}

\section{Introduction}
Dense prediction is a fundamental task in computer vision, benefiting a wide range of applications, such as 3D/4D reconstruction~\citep{Huang2DGS2024,long2024wonder3d,wang2024shape,lei2024mosca}, tracking~\citep{SpatialTracker,song2024track}, and autonomous driving~\citep{yurtsever2020survey, hu2023planning}.
\haodong{Estimating pixel-level geometric attributes from a single image requires comprehensive scene understanding. Although deep learning has advanced dense prediction, progress is limited by the quality, diversity, and scale of training data, leading to poor zero-shot generalization.}
\haodong{Instead of merely scaling data and model size, recent works~\citep{lee2024exploiting, ke2024repurposing, fu2024geowizard, xu2024diffusion} leverage diffusion priors for zero-shot dense prediction. These studies demonstrate that text-to-image diffusion models like Stable Diffusion~\citep{rombach2022high}, pretrained on billions of images,}
possess powerful and comprehensive visual priors to elevate dense prediction performance. 
However, most of these methods directly inherit the pre-trained diffusion models for dense prediction tasks, without exploring more suitable diffusion formulations. This oversight often leads to challenging issues.
\haodong{For example, Marigold~\citep{ke2024repurposing} directly fine-tunes Stable Diffusion for image-conditioned depth generation. While it significantly improves depth estimation, its performance is still constrained by overlooking the fundamental differences between dense prediction and image generation. Especially, its efficiency is also severely limited by standard iterative denoising processes and ensemble inferences.}

\begin{figure}[t]
    \centering
    \includegraphics[width = 1.0\linewidth]{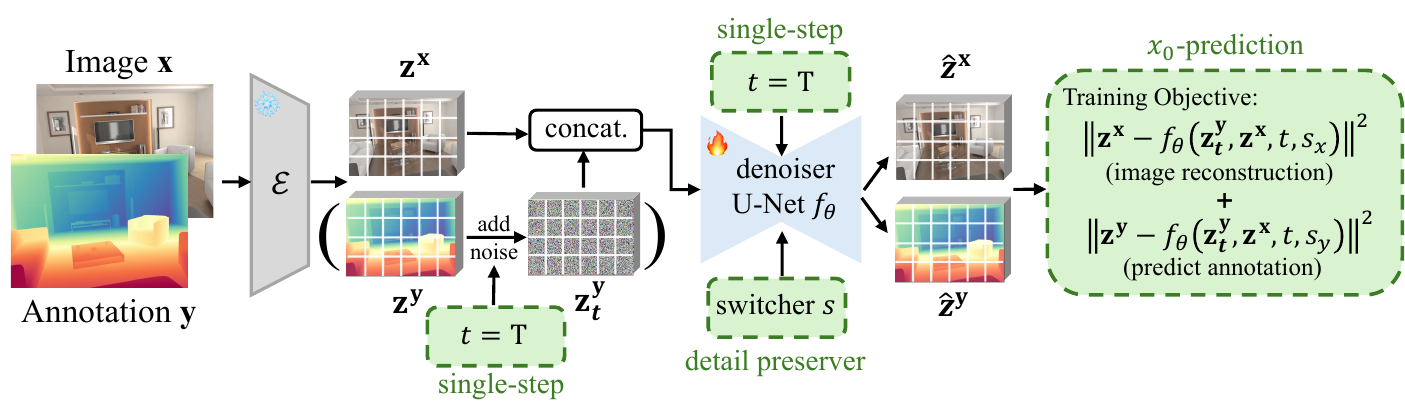}
    \caption{\textbf{Adaptation protocol of Lotus.} 
    After the pre-trained VAE encoder $\mathcal{E}$ encodes the image $\textbf{x}$ and annotation $\textbf{y}$ to the latent space: \ding{172} the denoiser U-Net model $f_\theta$ is fine-tuned using $x_0$-prediction; \ding{173} we employ single-step diffusion formulation at time-step $t=T$ for better convergence; \ding{174} we propose a novel detail preserver, to switch the model either to reconstruct the image or generate the dense prediction via a switcher $s$, ensuring a more fine-grained prediction.
    The noise $\mathbf{z_T^y}$ in bracket is used for our generative \textbf{Lotus-G} and is omitted for the discriminative \textbf{Lotus-D}.
    }
    \label{fig:training}
\end{figure}
Motivated by these concerns, we systematically analyze the diffusion formulation, trying to find a \haodong{better} formulation to fit the pre-trained diffusion model into dense prediction. 
Our analysis yields several important findings:
\ding{172} The widely used parameterization, \emph{i.e.}, noise prediction, for diffusion-based image generation is ill-suited for dense prediction. It results in large prediction errors due to harmful prediction variance at initial denoising steps, which are subsequently propagated and magnified throughout the entire denoising process (Sec.~\ref{sec:param}).
\ding{173} Multi-step diffusion formulation is computation-intensive and is prone to sub-optimal with limited data and resources. These factors significantly hinder the adaptation of diffusion priors to dense prediction tasks, leading to decreased accuracy and efficiency (Sec.~\ref{sec:step}).
\ding{174} Though remarkable performance achieved, we observed that the model usually outputs vague predictions in highly-detailed areas (Fig.~\ref{fig:enhancer}). 
\haodong{This vagueness is attributed to catastrophic forgetting: the pre-trained diffusion models gradually lose their ability to generate detailed regions during fine-tuning (Sec.~\ref{sec:DE}).}

\begin{wrapfigure}{r}{0.49\textwidth}
    \centering
    \includegraphics[width = 1.0\linewidth]{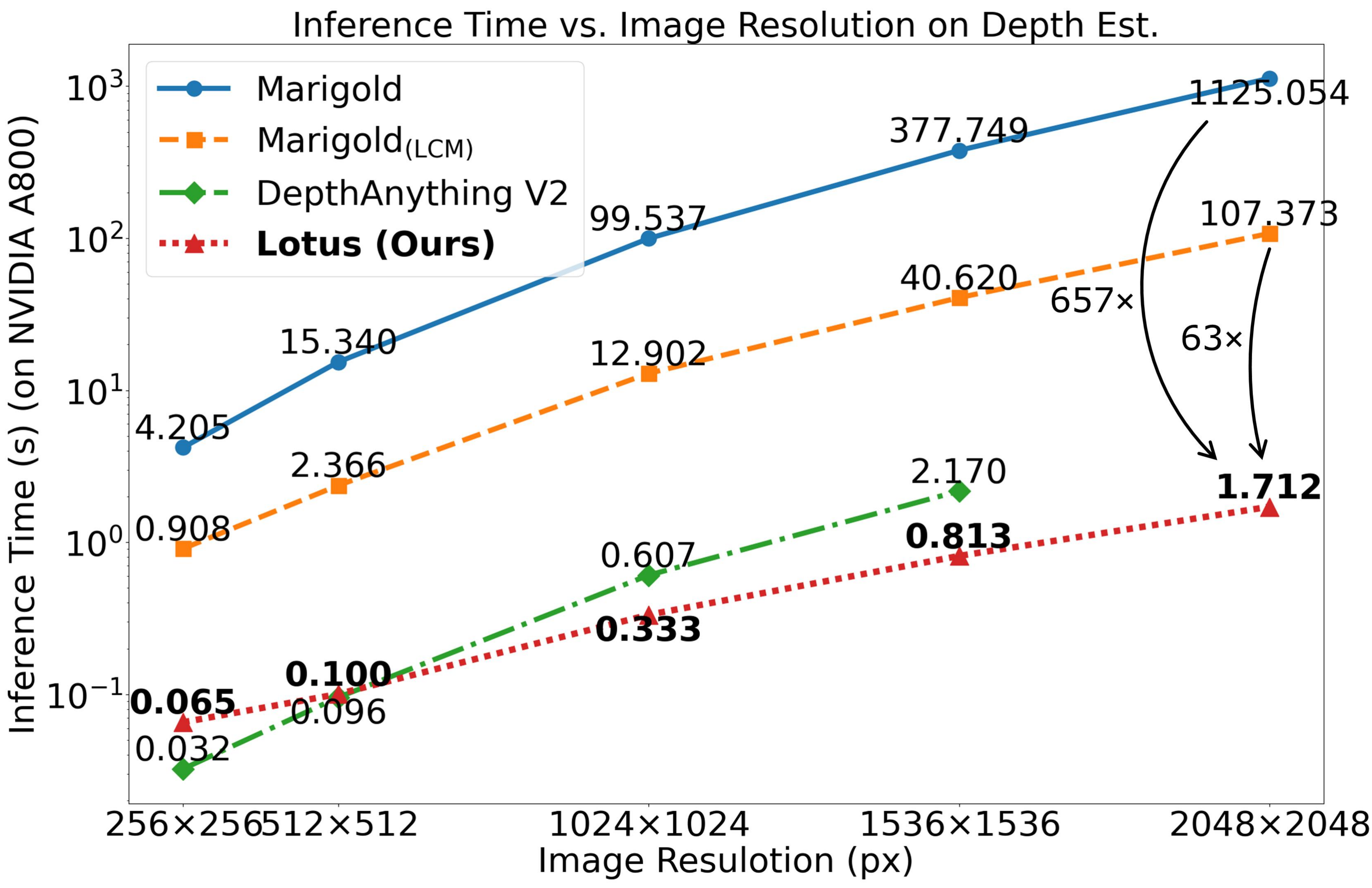}
    \caption{\textbf{Inference time comparison in depth estimation between Lotus and SoTA methods.} Lotus is hundreds of times faster than Marigold and slightly faster than DepthAnything V2 at high resolutions.
    \haodong{DepthAnything V2's inference time at $2048\times 2048$ is not plotted because it requires $>80$GB graphic memory.}}
    \label{fig:speed}
\end{wrapfigure}
Following our analysis, we propose \textbf{Lotus}, a diffusion-based visual foundation model for dense prediction, featuring a simple yet effective fine-tuning protocol (see Fig.~\ref{fig:training}).
First, Lotus is trained to directly predict annotations, thereby avoiding the harmful variance associated with standard noise prediction. Next, we introduce a one-step formulation, \emph{i.e.}, one step between pure noise and clean output,
to facilitate model convergence and achieve better optimization performance with limited high-quality data. \haodong{It also considerably boosts both training and inference efficiency.
Moreover, we implement a novel detail preserver through a task switcher, allowing the model either to generate annotations or reconstruct the input images.}
It can better preserve the fine-grained details in  input image during dense annotation generation, achieving higher performance without compromising efficiency, requiring additional parameters, or being affected by surface textures.

To validate Lotus, we conduct extensive experiments on two primary geometric dense prediction tasks: zero-shot monocular depth and normal estimation. The results demonstrate that Lotus achieves promising, and even superior, performance on these tasks across a wide range of evaluation datasets. 
Compared to traditional discriminative methods, Lotus delivers remarkable results with only 59K training samples by effectively leveraging the powerful diffusion priors.
Among generative approaches, Lotus also outperforms previous methods in both accuracy and efficiency, being significantly faster than methods like Marigold~\citep{ke2024repurposing}
(Fig.~\ref{fig:speed}).
Beyond these improvements, Lotus seamlessly supports various applications, such as joint estimation, single/multi-view 3D reconstruction, etc.

In conclusion, our key contributions are as follows: 
\begin{itemize}
    \item We systematically analyze the diffusion formulation and find their parameterization type, designed for image generation, is unsuitable for dense prediction and the computation-intensive multi-step diffusion process is also unnecessary and challenging to optimize.
    \item We propose a novel detail preserver that ensures more accurate dense predictions especially in detail-rich areas, \haodong{without compromising efficiency, introducing additional network parameters, or being affected by surface textures.}
    \item Based on our insights, we introduce \textbf{Lotus}, a diffusion-based visual foundation model for dense prediction with simple yet effective fine-tuning protocol. 
    Lotus achieves promising performance on both zero-shot monocular depth and surface normal estimation. It also enables a wide range of applications. 
\end{itemize}

\section{Related Works}
\subsection{Text-to-Image Generative Models}
In the field of text-to-image generation, the evolution of methodologies has transitioned from generative adversarial networks (GANs)~\citep{goodfellow2014generative, zhang2017stackgan, zhang2018stackgan++, zhang2021cross, pixelfolder, StyleGAN1, StyleGAN2, StyleGAN3,zhang2017stackgan, zhang2018stackgan++,xu2018attngan,zhang2021cross} to advanced diffusion models~\citep{ho2020denoising,unclip,imagen,dalle,nichol2021glide,pixart, rombach2022high,dalle,he2024disenvisioner}.
A series of diffusion-based methods such as GLIDE~\citep{nichol2021glide}, DALL$\cdot$E2~\citep{unclip}, and Imagen~\citep{imagen} have been introduced, offering enhanced image quality and textual coherence. 
The Stable Diffusion (SD)~\citep{rombach2022high},
trained on large-scale LAION-5B dataset~\citep{schuhmann2022laion}, further enhances the generative quality, becoming the community standard.
In our paper, we aim to leverage the comprehensive and encyclopedic visual priors of SD to facilitate zero-shot generalization for dense prediction tasks.

\subsection{Generative Models for Dense Perception}
Currently, a notable trend involves adopting pre-trained generative models, particularly diffusion models, into dense prediction tasks. 
\haodong{Marigold~\citep{ke2024repurposing} and GeoWizard~\citep{fu2024geowizard} directly apply the standard diffusion formulation and the pre-trained parameters, without addressing the inherent differences between image generation and dense prediction, leading to constrained performance.
Their efficiency is also severely limited by standard iterative denoising processes and ensemble inferences. 
In this paper, we propose a novel diffusion formulation tailored to the of dense prediction. Aiming to fully leveraging the pre-trained diffusion's powerful visual priors, Lotus enables more accurate and efficient predictions, finally achieving promising performance.}

More recent works, GenPercept~\citep{xu2024diffusion} and StableNormal~\citep{ye2024stablenormal}, also adopted single-step diffusion. However, GenPercept~\citep{xu2024diffusion} first removes noise input for deterministic characteristic based on DMP~\citep{lee2024exploiting}, and then adopts one-step strategy to avoid surface texture interference. It lacks systematic analysis of the diffusion formulation, only treats the U-Net as a deterministic backbone and still falls short in performance.
In contrast, Lotus systematically analyzes the standard stochastic diffusion formulation for dense prediction and proposes innovations such as the detail preserver to improve accuracy especially in detailed area, finally delivering much better results (Tab.~\ref{tab:depth}). Additionally, 
Lotus is a stochastic model. In contrast to GenPercept's deterministic nature, Lotus enables uncertainty predictions.
StableNormal~\citep{ye2024stablenormal} predicts normal maps through a two-stage process.
While the first stage produces coarse normal maps with single-step diffusion, the second stage performs refinement still with iterative diffusion which is computation-intensive. 
In comparison, Lotus not only achieves fine-grained predictions thanks to our novel detail preserver without extra stages or parameters, but also delivers much superior results (Tab.~\ref{tab:normal}) thanks to our designed diffusion formulation that better fits the pre-trained diffusion for dense prediction.
Recently, a concurrent work, Diffusion-E2E-FT \citep{garcia2024fine}, has also achieved promising results in a single step.
Its main contribution lies in addressing the issue where Marigold~\citep{ke2024repurposing} and similar models~\citep{fu2024geowizard} use inconsistent pairings of time-step and noise, resulting in poor predictions.  By setting the ``time-step spacing'' to ``trailing'' mode in schedulers, it prevents ``GT'' signal leakage during inference, improving accuracy.
While the performance of Lotus-D and Diffusion-E2E-FT is similar, Lotus is based on a systematic analysis of stochastic diffusion for dense prediction, with innovations like the detail preserver to enhance accuracy, particularly in detailed areas. Additionally, unlike the deterministic Diffusion-E2E-FT, Lotus (Lotus-G) is a stochastic model that enables uncertainty predictions.

\subsection{Monocular Depth and Normal Prediction}
Monocular depth and normal prediction are two crucial dense prediction tasks. Solving them typically demands comprehensive scene understanding capability.
Starting from \citep{eigen2014depth}, early CNN-based methods for depth prediction, such as \citep{fu2018deep,lee2019big,yuan2022neural}, focus only on specific domains. 
Subsequently, in pursuit of a generalizable depth estimator, many methods expand model capacity and train on larger and more diverse datasets, such as DiverseDepth~\citep{yin2021virtual} and MiDaS~\citep{ranftl2020towards}. 
DPT~\citep{ranftl2021vision} and Omnidata~\citep{eftekhar2021omnidata} are further proposed based on vision transformer~\citep{ranftl2021vision}, significantly enhancing performance. LeRes~\citep{yin2021learning} and HDN~\citep{zhang2022hierarchical} further introduce novel training strategies and multi-scale depth normalization to improve predictions in detailed areas. More recently, the DepthAnything series~\citep{yang2024depth1, yang2024depth2} and Metric3D series~\citep{yin2023metric3d, hu2024metric3d} collect and leverage millions of training data to develop more powerful estimators. 
Normal prediction follows the same trend. Starting with the early CNN-based methods like OASIS~\citep{chen2020oasis}, EESNU~\citep{Bae2021} and Omnidata series~\citep{eftekhar2021omnidata,kar20223d} expand the model capacity and scale up the training data. Recently, DSINE~\citep{bae2024dsine} achieves SoTA performance by rethinking inductive biases for surface normal estimation. 
In our paper, we focus on leveraging pre-trained diffusion priors to enhance zero-shot dense predictions, rather than expanding model capacity or relying on large training data, which avoids the need for intensive resources and computation. 

\section{Preliminaries}
\label{sec:pre}
\textbf{Diffusion Formulation for Dense Prediction.} 
Following \citep{ke2024repurposing} and \citep{fu2024geowizard}, we also formulate dense prediction as an image-conditioned annotation generation task based on Stable Diffusion~\citep{rombach2022high}, which performs the diffusion process in low-dimensional latent space for computational efficiency. First, the auto-encoder, which consists an encoder $\mathcal{E}(\cdot)$ and a decoder $\mathcal{D}(\cdot)$, is trained to map between RGB space and latent space, \emph{i.e.}, $\mathcal{E}(\textbf{x})=\mathbf{z^x}$, $\mathcal{D}(\mathbf{z^x})\approx \textbf{x}$. The auto-encoder also maps between dense annotations and latent space effectively, \emph{i.e.}, $\mathcal{E}(\textbf{y})=\mathbf{z^y}$, $\mathcal{D}(\mathbf{z^y})\approx \textbf{y}$~\citep{ke2024repurposing, fu2024geowizard,xu2024diffusion,ye2024stablenormal}.
Following \citep{ho2020denoising}, Stable Diffusion establishes a pair of \emph{forward} nosing and \emph{reversal} denoising processes in latent space. In \emph{forward} process, Gaussian noise is gradually added at levels $t\in \left[1,T\right]$ into sample $\mathbf{z^y}$ to obtain the noisy sample $\mathbf{z^y_t}$: 
\begin{equation}
\label{eq:xt}
    \mathbf{z^y_t} = \sqrt{\overline{\alpha}_t}\mathbf{z^y} + \sqrt{1-\overline{\alpha}_t}\mathbf{\epsilon}, 
\end{equation}
where $\epsilon \sim \mathcal{N}(0, I)$, $\overline{\alpha}_t := \prod_{s=1}^t (1- \beta_s)$, and $\{\beta_1, \beta_2, \dots, \beta_T\}$ is the noise schedule with $T$ steps. At time-step $T$, the sample $\mathbf{z^y}$ is degraded to pure Gaussian noise. In the \emph{reversal} process, a neural network $f_\theta$, usually a U-Net model \citep{ronneberger2015u}, is trained to iteratively remove noise from $\mathbf{z^y_t}$ to predict the clean sample $\mathbf{z^y}$. The network is trained by sampling a random $t \in \left[1, T\right]$ and minimizing the loss function $L_t$.

\textbf{Parameterization Types.} 
To enable gradient computation for network training, there are two basic parameterizations of the loss function $L_t$. 
\ding{172} $\epsilon$-prediction \citep{ho2020denoising}: the model $f_{\theta}$ learns to predict the added noise $\epsilon$;
\ding{173} $x_0$-prediction \citep{ho2020denoising}: the model $f_{\theta}$ learns to directly predict the clean sample $\mathbf{z^y}$.
The loss functions for these parameterizations are formulated as:
\begin{equation}
\label{eq:eps}
\begin{aligned}
    \epsilon\text{-prediction: }&L_t^\epsilon = ||\epsilon-f_\theta^\epsilon(\mathbf{z^y_t}, \mathbf{z^x}, t)||^2,\\
    x_0\text{-prediction: }&L_t^\textbf{z} = ||\mathbf{z^y}-f_\theta^\textbf{z}(\mathbf{z^y_t}, \mathbf{z^x}, t)||^2. 
\end{aligned}
\end{equation}
where $f_\theta^{*}$ is the denoiser model to be learnt, ${*} \in \{\epsilon,\textbf{z}\}$. $\epsilon$-prediction is commonly chosen as the standard for parameterizing the denoising model, as it empirically achieves high-quality image generation with fine details and realism.

\begin{figure}[t]
    \centering
    \includegraphics[width = 1.0\linewidth]{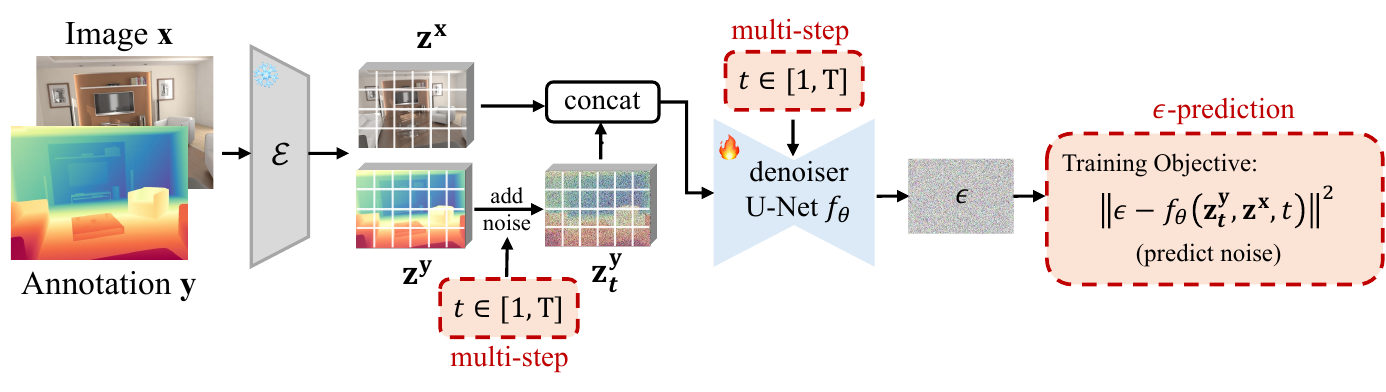}
    \caption{\textbf{Adaptation protocol of Direct Adaptation.} Starting with a pre-trained Stable Diffusion model, image $\textbf{x}$ and annotation $\textbf{y}$ are encoded using the pre-trained VAE. The noisy annotation $\mathbf{z^y_{t}}$ is obtained by adding noise at level $t \in \left[1, T\right]$. The U-Net input layer is coupled to accommodate the concatenated inputs and then fine-tuned using the standard diffusion objective, $\epsilon$-prediction, under the original multi-step formulation.
    }
    \label{fig:dada}
\end{figure}

\textbf{Denoising Process. }
DDIM~\citep{song2020denoising} is a key technique for multi-step diffusion models to achieve fast sampling,
which implements an implicit probabilistic model that can significantly reduce the number of denoising steps while maintaining output quality. Formally, the denoising process from $\mathbf{z^y_\tau}$ to $\mathbf{z^y_{\tau-1}}$ is: 
\begin{equation}
\label{eq:ddim}
    \mathbf{z^y_{\tau-1}} = \sqrt{\overline{\alpha}_{\tau-1}} \mathbf{\hat{z}^y_\tau} + \text{direction}(\mathbf{z^y_{\tau}}) + \sigma_\tau\epsilon_\tau, 
\end{equation}
where $\mathbf{\hat{z}^y_\tau}$ is the predicted clean sample at the denoising step $\tau$, $\text{direction}(\mathbf{z^y_{\tau}})$ represents the direction pointing to $\mathbf{z^y_{\tau}}$ and $\sigma_\tau$ can be set to $0$ if deterministic denoising is needed. And $\tau \in \{ \tau_1, \tau_2, \dots, \tau_S \}$, an increasing sub-sequence of the time-step set $[1,T]$, is used for fast sampling. During inference, DDIM iteratively denoises the sample from $\tau_S$ to $\tau_1$ to obtain the clean one. 

\section{Methodology}
\label{Sec: method}
We start our analysis by directly adapting the original diffusion formulation with minimal modifications as illustrated in Fig.~\ref{fig:dada}. We call this starting point as ``\emph{Direct Adaptation}''\footnote{Details of Direct Adaptation will be provided in Sec.~\ref{suppl:da} of the supplementary materials.}.
Direct Adaptation is optimized using the standard diffusion objective as formulated in Eq.~\ref{eq:eps} (first row) and inferred by standard multi-step DDIM sampler. 
As shown in Tab.~\ref{tab:ablation}, Direct Adaptation fails to achieve satisfactory performance.
\haodong{In following sections, we will systematically analyze the key factors that affect adaptation performance step by step: parameterization types (Sec.~\ref{sec:param}); number of time-steps (Sec.~\ref{sec:step}); and the novel detail preserver (Sec.~\ref{sec:DE}).}

\subsection{Parameterization Types}
\label{sec:param}
The type of parameterization is crucial, it not only determines the loss function discussed in Sec.~\ref{sec:pre}, but also influences the inference process (Eq.~\ref{eq:ddim}). 
During inference, the predicted clean sample $\mathbf{\hat{z}^y_\tau}$, a key component in Eq.~\ref{eq:ddim}, is calculated according to different parameterizations~\footnote{The latest parameterization, $v$-prediction, combines $\epsilon$-prediction and $x_0$-prediction, producing results that are intermediate between the two. Please see Sec.~\ref{suppl:v-pred} of the supplementary materials for more details.}.
\begin{equation}
\label{eq:param_types}
\begin{aligned}
\epsilon\text{-prediction: }\mathbf{\hat{z}^y_\tau} &= \frac{1}{\sqrt{\overline{\alpha}_\tau}}(\mathbf{z^y_\tau} - \sqrt{1-\overline{\alpha}_\tau}f_\theta^\epsilon(\mathbf{z^y_\tau}, \mathbf{z^x}, \tau)),\\
x_0\text{-prediction: }\mathbf{\hat{z}^y_\tau} &= f_\theta^\textbf{z}(\mathbf{z^y_\tau}, \mathbf{z^x}, \tau).
\end{aligned}
\end{equation}

In the community, $\epsilon$-prediction is chosen as the standard for image generation. However, it is not effective for dense prediction task. 
In the following, we will discuss the impact of different parameterization types in denoising inference process for dense prediction task. 

\begin{figure}[t]
    \centering
    \includegraphics[width = 1.0\linewidth]{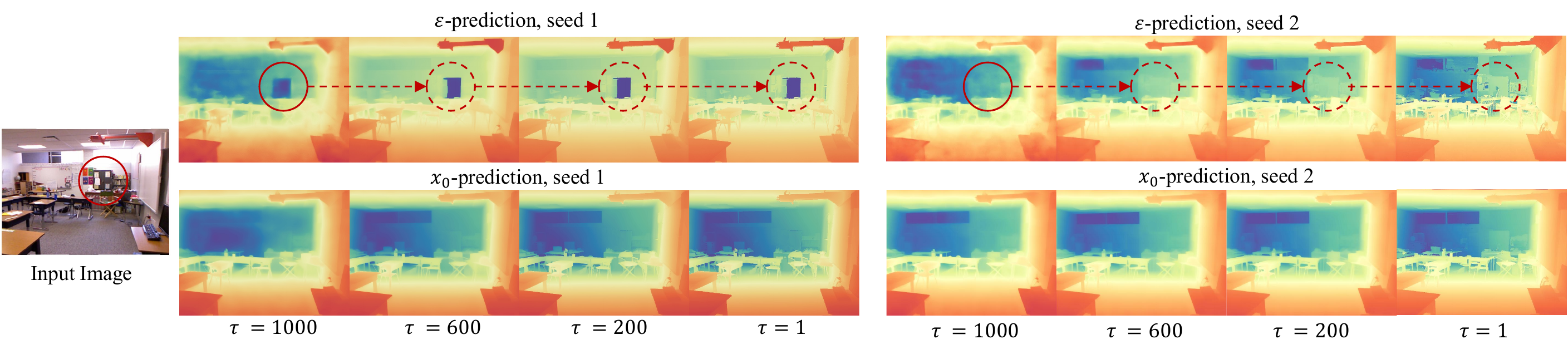}
    \caption{\textbf{Comparisons among different parameterizations using various seeds.} All models are trained on \textit{Hypersim}~\citep{roberts2021hypersim} and tested on the input image for depth estimation. 
    The standard DDIM sampler is used with 50 denoising steps. Four steps are selected for clear illustration. From left (larger $\tau$) to right (smaller $\tau$) is the iterative denoising process.
    }
    \label{fig:3param}
\end{figure}

Insights from the literature~\citep{benny2022dynamic, salimans2022progressive} reveal that $\epsilon$-prediction introduces larger pixel variance compared to $x_0$-prediction, especially at the initial denoising steps (large $\tau$).
\hd{This variance mainly originates from the noise input.}
Specifically,
for $\epsilon$-prediction in Eq.~\ref{eq:param_types}, at initial denoising step, $\tau\rightarrow T$, the value $\frac{1}{\sqrt{\overline{\alpha}_\tau}}\rightarrow +\infty$. \hd{Thus, the prediction variance from $f_\theta^\epsilon(\mathbf{z^y_\tau}, \mathbf{z^x}, \tau)$ will be amplified significantly}, resulting in large variance of predicted $\mathbf{\hat{z}^y_\tau}$.
In contrast, there is no coefficient for $x_0$-prediction to re-scale the model output, achieving more stable predictions of $\mathbf{\hat{z}^y_\tau}$ at initial denoising steps. 
Subsequently, the predicted $\mathbf{\hat{z}^y_\tau}$ is used in Eq.~\ref{eq:ddim}, where its coefficient $\sqrt{\overline{\alpha}_{\tau-1}}$ are same across the two parameterizations, and other terms are of the same order of magnitude. Therefore, the $\mathbf{\hat{z}^y_\tau}$ predicted by $\epsilon$-prediction, which has larger variance, exerts a more significant influence on denoising process. Since the process is iterative, this influence is continually preserved and maybe amplified. 

\begin{wrapfigure}{t}{0.5\textwidth}
    \centering
    \includegraphics[width = \linewidth]{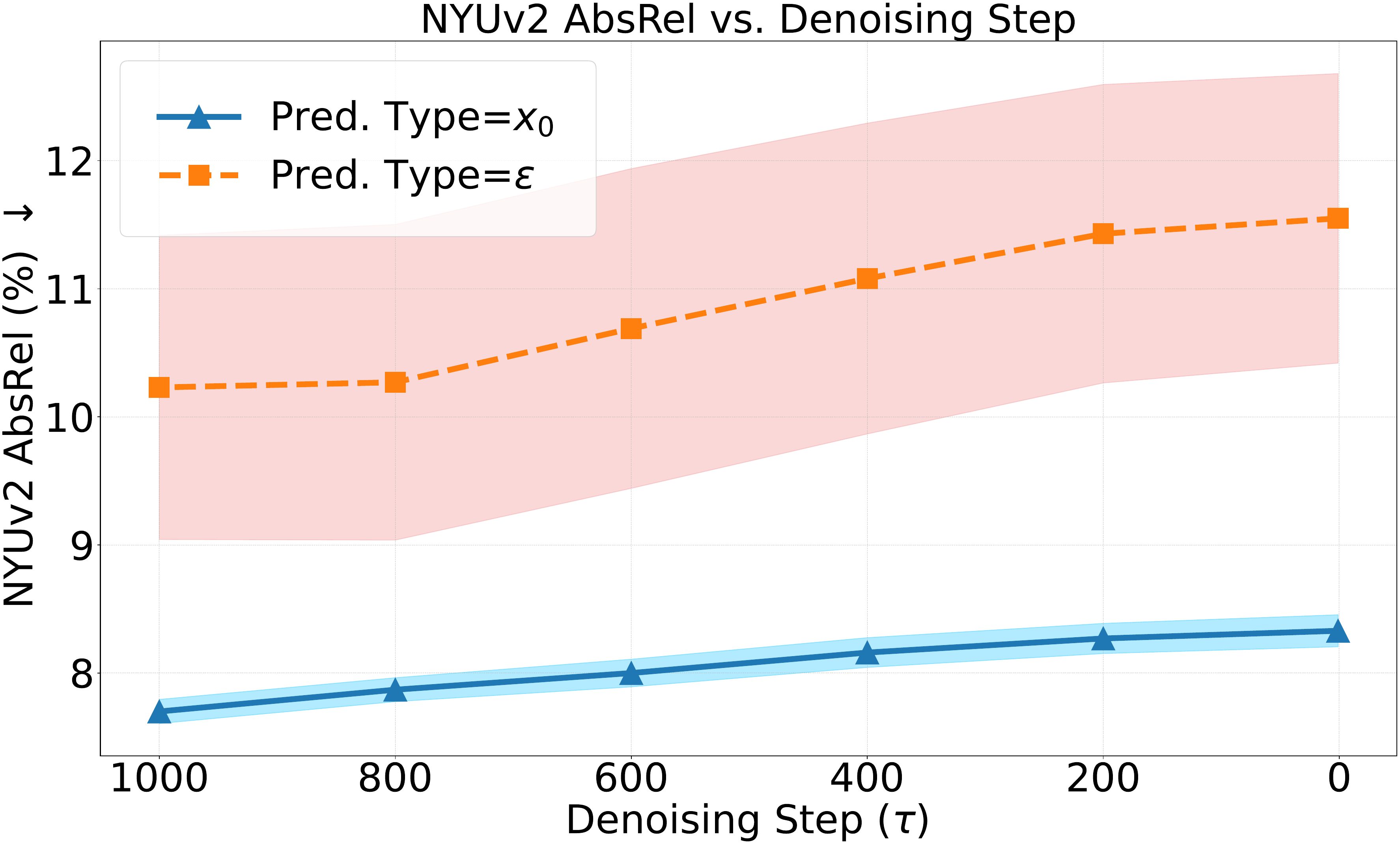}
    \caption{\textbf{Quantitative evaluation of the predicted depth maps $\mathbf{\hat{z}^y_\tau}$ along the denoising process.} The experimental settings are same as Fig.~\ref{fig:3param}. Six steps are selected for illustration. The banded regions around each line indicate the variance, wider areas representing larger variance. 
    }
\label{fig:pixe_error_with_variance}
\end{wrapfigure}
We take the depth estimation as an example.
During the inference process, we compute the predicted depth map $\mathbf{\hat{z}^y_\tau}$ at each denoising step $\tau$. 
As illustrated in Fig.~\ref{fig:3param}, the depth maps predicted by $\epsilon$-prediction significantly vary under different seeds while those predicted by $x_0$-prediction are more consistent.
Although the large variance enhances diversity for image generation, it lead to unstable predictions in dense prediction tasks, potentially resulting in significant errors. For example in Fig.~\ref{fig:3param}, the ``dark gray cabinet'' (highlighted in red circles) \haodong{maybe} wrongly considered as an “opened door” with significantly larger depth. While the predicted depth map \textit{looks} more and more \textit{plausible}, the error gradually propagates to the final prediction ($\tau=1$) along the denoising process, indicating the persistent influence of the large variance. We further quantitatively measure the predicted depth maps by the absolute mean relative error (AbsRel) on NYUv2 dataset~\citep{silberman2012indoor}.
As shown in Fig.~\ref{fig:pixe_error_with_variance}, $\epsilon$-prediction exhibits higher error with \haodong{much} larger variance compared to $x_0$-prediction at the initial denoising steps ($\tau\rightarrow T$), and the prediction error propagates with a higher slope. 
In contrast, $x_0$-prediction, directly predicting $\mathbf{\hat{z}^y_\tau}$ without any coefficients to amplify the prediction variance, yields more stable and correct dense predictions than $\epsilon$-prediction. In conclusion, to mitigate the errors from large variance that adversely affect the performance of dense prediction, we replace the standard $\epsilon$-prediction with the more tailored $x_0$-prediction.

\subsection{Number of Time-Steps}
\label{sec:step}
Although $x_0$-prediction can improve the prediction quality, the multi-step diffusion formulation still leads to the propagation of predicted errors during the denoising process (Fig.~\ref{fig:3param},~\ref{fig:pixe_error_with_variance}). 
\haodong{
Furthermore, utilizing multiple time-steps enhances the model's capacity, typically requiring large-scale training data to optimize and is beneficial for complex tasks such as image generation. However, for simpler tasks like dense prediction, where large-scale, high-quality training data is also scarce, employing multiple time-steps can make the model difficult to optimize.
Additionally, training/inferring a multi-step diffusion model is slow and computation-intensive, hindering its practical application.}

Therefore, to address these challenges, we propose fine-tuning the pre-trained diffusion model with fewer training time steps.
Specifically, the original set of training time-steps is defined as $[1,T] = \{1, 2, 3, \dots, T\}$, where $T$ denotes the total number of original training time-steps. We fine-tune the pre-trained diffusion model using a sub-sequence derived from this set. 
We define the length of this sub-sequence as $T'$, where $T' \leqslant T$ and $T$ is divisible by $T'$.
This sub-sequence is obtained by evenly sampling the original set at intervals, defined as:
\begin{equation}
    \{t_i = i \cdot k \mid i = 1, 2, \dots, T'\},
\label{eq:subset}
\end{equation}
where $k = T/T'$ is the sampling interval. During inference, the DDIM denoises the sample from noise to annotation using the same sub-sequence if $T'\leqslant50$, otherwise we use $50$ denoising steps.

\begin{wrapfigure}{t}{0.5\textwidth}
    \centering
    \includegraphics[width = 1.0\linewidth]{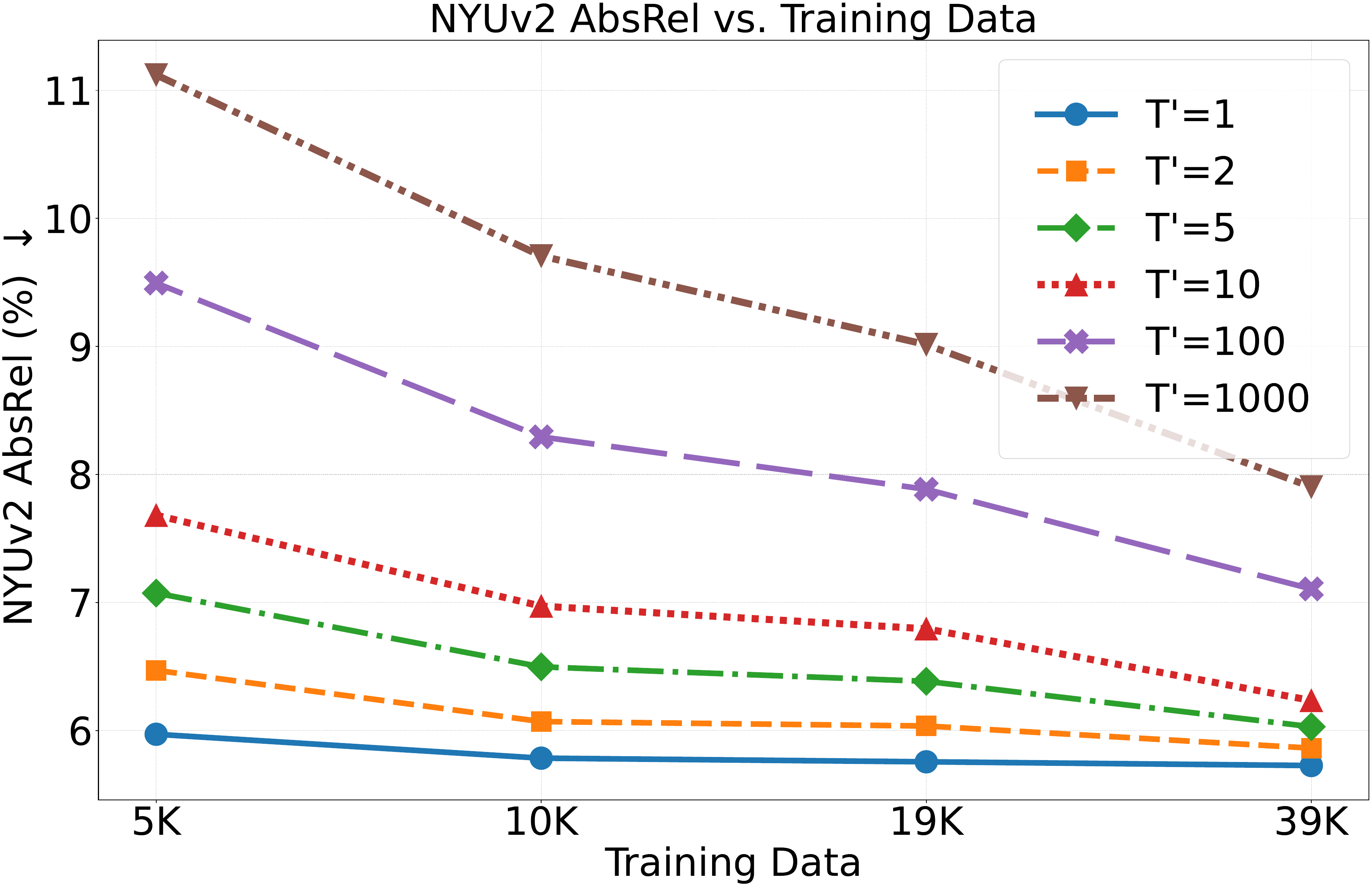}
    \caption{\textbf{Comparisons among various training time-steps and data scales} evaluated on NYUv2 in depth estimation. All models are fine-tuned on \textit{Hypersim} using $x_0$-prediction. 
    \haodong{During inference,} if $T'>50$, the DDIM sampler is used with 50 denoising steps; otherwise, the number of denoising steps is equal to $T'$. 
    The results demonstrate improved performance with decreased training time-steps. The single-step diffusion formulation ($T'=1$) exhibits best performance across different data volumes. 
    }
\label{fig:T}
\end{wrapfigure}

As illustrated in Fig.~\ref{fig:T}, we conduct experiments by varying the number of time-steps $T'$ under $x_0$-prediction. 
The results clearly show that the performance gradually improves as the number of time-steps is reduced, no matter the training data scales, culminating in the best result when reduced to only a single step. 
We further consider more strict scenarios with \haodong{more} limited training 
 data to assess its impact on model optimization. 
As depicted in Fig.~\ref{fig:T}, these experiments reveal that the multi-step formulation is more sensitive to increases in training data scales compared with single-step.
Notably, the single-step formulation consistently yields lower prediction errors and demonstrates greater stability. Although it is conceivable that multi-step and single-step formulations might achieve comparable performance with unlimited high-quality data, it's expensive and sometimes impractical in dense prediction.

Decreasing the number of denoising steps can reduce the optimization space of the diffusion model, leading to more effective and efficient adaption, as suggested by the above phenomenon.
Therefore, for better adaptation performance under limited resource, we reduce the number of training time-steps of diffusion formulation to only one, and fixing the only time-step $t$ to $T$. 
Additionally, the single-step formulation is much more computationally efficient. It also naturally prevents the harmful error propagation as discussed in Sec.~\ref{sec:param}, further enhancing the diffusion's adaptation performance in dense prediction.

\begin{figure}[t]
    \centering
    \includegraphics[width = 1.0\linewidth]{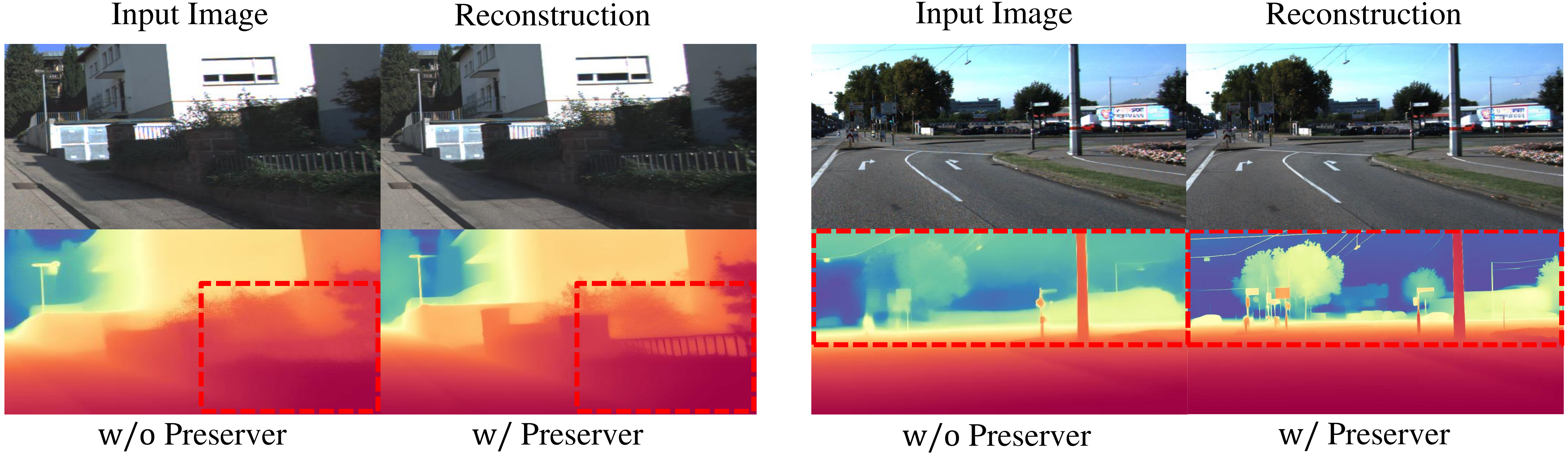}
    \caption{\textbf{Depth maps $w/$ and $w/o$ the detail preserver and reconstruction outputs.} Fine-tuning the diffusion model for dense prediction tasks can potentially degrade its ability to generate highly detailed images, resulting in blurred predictions in regions with rich detail. To preserve these fine-grained details, we introduce a detail preserver that incorporates an additional reconstruction task, enhancing the model's capacity to produce more accurate dense annotations.}  
    \label{fig:enhancer}
\end{figure}

\subsection{Detail Preserver}
\label{sec:DE}
Despite the effectiveness of the above designs, the model still struggles with processing detailed areas (Fig.~\ref{fig:enhancer}, $w/o$ Preserver).
The original diffusion model excels at generating detailed images. \haodong{However, when adapted to predict dense annotations, it can lose such detailed generation ability, 
due to unexpected catastrophic forgetting~\citep{zhai2023investigating,du2024unlocking}. This leads to challenges in predicting dense annotations in intricate regions.}

To preserve the rich details of the input images, we introduce a novel regularization strategy called \emph{Detail Preserver}. 
Inspired by previous works~\citep{long2024wonder3d,fu2024geowizard}, we utilize a task switcher $s\in\{s_x, s_y\}$, enabling the denoiser model $f_\theta$ to either generate annotation or reconstruct the input image. 
When activated by $s_y$, the model focuses on predicting annotation. Conversely, when $s_x$ is selected, it reconstructs the input image. 
The switcher $s$ is a one-dimensional vector encoded by the positional encoder and then \haodong{added}
with the time embeddings of diffusion model,
ensuring seamless domain switching without mutual interference.
\haodong{This dual capability enables the diffusion model to make detailed predictions and thus leading to better performance.}
Overall, the loss function $L_t$ is:
\begin{equation}
    L_t = ||\mathbf{z^x}-f_\theta(\mathbf{z_t^y}, \mathbf{z^x}, t, s_x)||^2 + ||\mathbf{z^y}-f_\theta(\mathbf{z_t^y}, \mathbf{z^x}, t, s_y)||^2,
\end{equation}
\haodong{where $t=T$ and thus $\mathbf{z_t^y}$ is a pure Gaussian noise. }

\begin{figure}[t]
    \centering
    \includegraphics[width = 1.0\linewidth]{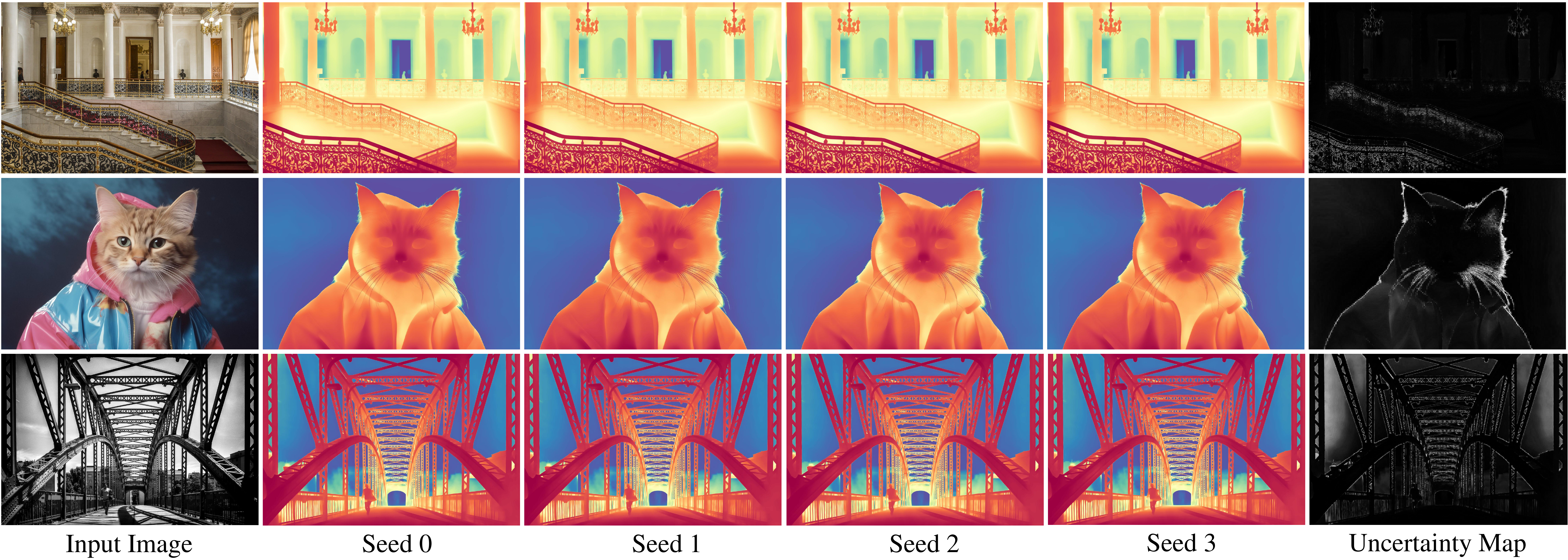}
    \caption{\textbf{Depth maps of multiple inferences and uncertainty maps.} Areas like the sky, object edges, and intricate details (\emph{e.g.}, cat whiskers) typically exhibit high uncertainty.
    }
    \label{fig:uncertainty}
\end{figure}
\subsection{Stochastic Nature of Diffusion Model}
One major characteristic of generative models is their stochastic nature, which, in image generation, enables the production of diverse outputs. \haodong{In perception tasks like dense prediction, this stochasticity has the potential to allow the model generating predictions with uncertainty maps.} Specifically, for any input image, we can conduct multiple inferences using different initialization noises and aggregate these predictions to calculate its uncertainty map.
Thanks to our systematic analysis and tailored fine-tuning protocol, our method effectively 
\begin{wrapfigure}{r}{0.45\textwidth}
    \vspace{-5mm}
    \centering
    \includegraphics[width = 1.0\linewidth]{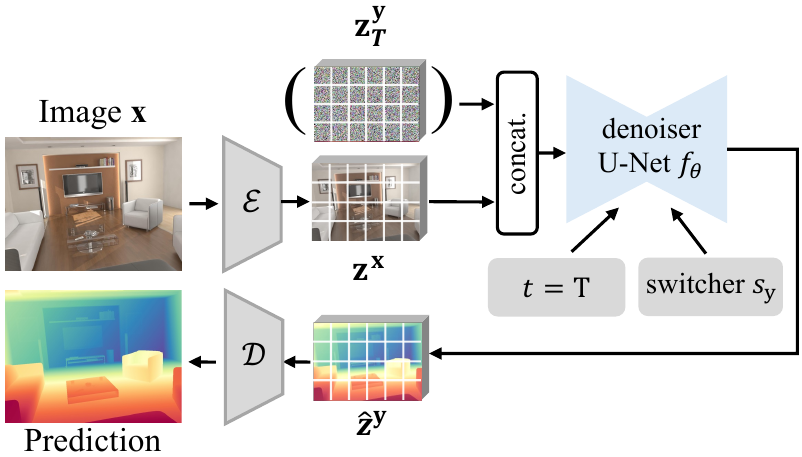}
    \caption{\textbf{Inference Pipeline of Lotus.} The noise $\mathbf{z_T^y}$ in bracket is used for \textbf{Lotus-G} and omitted for \textbf{Lotus-D}. 
    }
    \vspace{-5mm}
    \label{fig:inference}
\end{wrapfigure}
reduces excessive flickering (large variance), only allowing for more accurate uncertainty calculations in naturally uncertain areas, such as the sky, object edges, and fine details (\emph{e.g.} cat whiskers), as shown in Fig.~\ref{fig:uncertainty}. 

Most existing perception models are deterministic. To align with these, we can remove the noise input $\mathbf{z_t^y}$ and \haodong{only} \haodong{input the encoded image features $\mathbf{z^x}$ to the U-Net denoiser. The model still performs well.}
In this paper, we finally present two versions of Lotus: \textbf{Lotus-G} (generative) with noise input and \textbf{Lotus-D} (discriminative) without noise input, catering to different needs.

\subsection{Inference}
The inference pipeline is illustrated in Fig.~\ref{fig:inference}. 
We initialize the annotation map with standard Gaussian noise $\mathbf{z_T^y}$, and encode the input image into its latent code $\mathbf{z^x}$. The noise $\mathbf{z_T^y}$ and the image $\mathbf{z^x}$ are concatenated and fed into the denoiser U-Net model. In our single-step formulation, \haodong{we set $t=T$ and the switcher to $s_y$.} The denoiser U-Net model then predicts the latent code of the annotation map. The final annotation map is decoded from the predicted latent code via the VAE decoder. For deterministic prediction, we eliminate the Gaussian noise $\mathbf{z_T^y}$ and only feed the latent code of the input image into U-Net.

\section{Experiments}
\label{sec:exp}
\vspace{-2mm}
\subsection{Experimental Settings}
\vspace{-1mm}
\textbf{Implementation details.} We implement Lotus based on Stable Diffusion V2~\citep{rombach2022high}, without text conditioning.  During training, we fix the time-step $t=1000$. For depth estimation, we predict in disparity space, \textit{i.e.}, $d = 1/d'$, where $d$ represents the values in disparity space and $d'$ denotes the true depth. For more details, please see Sec.~\ref{suppl:impl} of the supplementary materials.

\textbf{Training Datasets.}
Both depth and normal estimation are trained on two synthetic dataset covering indoor and outdoor scenes: \ding{172} \emph{Hypersim}~\citep{roberts2021hypersim} is a photorealistic synthetic dataset featuring 461 indoor scenes. We use the official training split, which contains approximately 54K samples. After filtering out incomplete samples, around 39K samples remain, all resized to $576 \times 768$ for training. \ding{173} \emph{Virtual KITTI}~\citep{cabon2020virtual} is a synthetic street-scene dataset with five urban scenes under various imaging and weather conditions. We utilize four of these scenes for training, comprising about 20K samples. All samples are cropped to $352 \times 1216$, with the far plane at 80m.

Following Marigold~\citep{ke2024repurposing}, 
we probabilistically choose one of the two datasets and then draw samples from it for each batch (\emph{Hypersim} 90\% and \emph{Virtual KITTI} 10\%).

\textbf{Evaluation Datasets and Metrics.} 
\ding{172} For zero-shot affine-invariant depth estimation, we evaluate Lotus on NYUv2~\citep{silberman2012indoor}, ScanNet~\citep{dai2017scannet}, KITTI~\citep{geiger2013vision}, ETH3D~\citep{schops2017multi}, and DIODE~\citep{vasiljevic2019diode} using absolute mean relative error (\emph{AbsRel}), and also report $\delta 1$ and $\delta2$ values. \ding{173} For surface normal prediction, we employ NYUv2, ScanNet, iBims-1~\citep{koch2018evaluation}, Sintel~\citep{butler2012naturalistic}, and OASIS~\citep{chen2020oasis} datasets, reporting mean angular error (\emph{m.}) as well as the percentage of pixels with an angular error below $11.25^{\circ}$ and $30^\circ$.
Please see Sec.~\ref{suppl:eval_metrics} of the supplementary materials for further details on the evaluation datasets and metrics.

\vspace{-2mm}
\subsection{Quantitative Comparisons} 
\vspace{-1mm}
\ding{172} For depth estimation (Tab.~\ref{tab:depth}), Lotus-G demonstrates promising performance across all evaluation datasets, achieving the overall best rank compared to other generative baselines. Notice that we only require single step denoising process, significantly boosting the inference speed as shown in Fig.~\ref{fig:speed}. Lotus-D also performs well, achieving comparable results to DepthAnything series. It is worthy to notice that Lotus is trained on only 0.059M images compared to DepthAnything’s 62.6M images. 
\ding{173} For normal estimation (Tab.~\ref{tab:normal}), both Lotus-G and Lotus-D outperform all other generative and discriminative methods on zero-shot surface normal estimation with significant margins.
Please see Sec.~\ref{suppl:qualitative} of the supplementary materials for \textbf{Qualitative Comparisons}.

\vspace{-2mm}
\subsection{Ablation Study}
\vspace{-1mm}
As shown in Tab.~\ref{tab:ablation}, we conduct ablation studies to validate our designs. Starting with ``Direct Adaptation", we incrementally test the effects of different components, such as parameterization types, the single-step diffusion process, and the detail preserver. Initially, we train the model using only the \textit{Hypersim} dataset to establish a baseline. We then expand the training dataset using a mixture dataset strategy by including \textit{Virtual KITTI}, aiming to enhance the model's generalization ability across different domains. For depth estimation, we further train the model in the disparity space to improve the accuracy. The findings from these ablations validate the effectiveness of our proposed adaptation protocol, demonstrating that each design plays a vital role in optimizing the diffusion models for dense prediction tasks. 

\setlength{\tabcolsep}{2.pt}
\begin{table}[t]
\scriptsize
\caption{\textbf{Quantitative comparison on zero-shot affine-invariant depth estimation} between Lotus and SoTA methods. The upper section lists discriminative methods, the lower lists generative ones. The \colorbox{best}{best} and \colorbox{best2}{second best} performances are highlighted. \textbf{Lotus-G} outperforms all others methods while \textbf{Lotus-D} is only slightly inferior to DepthAnything.
$^\S$indicates results revised by ourselves, following Marigold~\citep{ke2024repurposing}. $^\star$denotes the method relies on pre-trained Stable Diffusion.}
\vspace{-3mm}
\label{tab:depth}
\scriptsize

\resizebox{\textwidth}{!}{
\begin{tabular}{l|c|ccc|ccc|ccc|ccc|ccc|c}
\toprule

\multirow{2}{*}{Method} & Training
& \multicolumn{3}{c|}{NYUv2 (Indoor)} & \multicolumn{3}{c|}{KITTI (Outdoor)} 
& \multicolumn{3}{c|}{ETH3D (Various)} & \multicolumn{3}{c|}{ScanNet (Indoor)} 
& \multicolumn{3}{c|}{\textcolor{black}{DIODE (Various)}} 
& \textcolor{black}{Avg}\\
 & Data$\downarrow$
 & AbsRel$\downarrow$ & $\delta$1$\uparrow$ & $\delta$2$\uparrow$ & AbsRel$\downarrow$ & $\delta$1$\uparrow$ & $\delta$2$\uparrow$
 & AbsRel$\downarrow$ & $\delta$1$\uparrow$ & $\delta$2$\uparrow$ & AbsRel$\downarrow$ & $\delta$1$\uparrow$ & $\delta$2$\uparrow$
 & \textcolor{black}{AbsRel$\downarrow$} & \textcolor{black}{$\delta$1$\uparrow$} & \textcolor{black}{$\delta$2$\uparrow$}
 &    \textcolor{black}{ Rank}        \\
\midrule

DiverseDepth 
 & 320K
& 11.7 & 87.5 & -& 19.0 & 70.4 & -    
& 22.8 & 69.4 & -& 10.9 & 88.2 & -
& \textcolor{black}{37.6} & \textcolor{black}{63.1} & -
& \textcolor{black}{10.6}               \\

MiDaS 
 &2M
& 11.1 & 88.5 & -&23.6 & 63.0 &-
& 18.4 & 75.2 & -&12.1 & 84.6 &-
& \textcolor{black}{33.2} & \textcolor{black}{71.5} & -
&  \textcolor{black}{10.2}                 \\

LeRes 
 &354K
& 9.0 & 91.6 &  -& 14.9 & 78.4 &-
& 17.1 & 77.7 & -& 9.1 & 91.7 & -
& \textcolor{black}{27.1} & \textcolor{black}{76.6} & -
& \textcolor{black}{7.8}                  \\

Omnidata
 &12.2M
& 7.4 & 94.5 & -& 14.9 & 83.5 &-
& 16.6 & 77.8 & -& 7.5 & 93.6 &-
& \textcolor{black}{33.9} & \textcolor{black}{74.2} & -
& \textcolor{black}{7.5} \\

DPT
 &1.4M
& 9.8 & 90.3 & -& \cellcolor{white}10.0 & \cellcolor{white}90.1 &-
& 7.8 & 94.6 & -& 8.2 & 93.4 &-
& \textcolor{black}{\cellcolor{best}18.2} & \textcolor{black}{75.8} & -
&  \textcolor{black}{5.8}                \\

HDN
 &300K
& 6.9 & 94.8 & -& 11.5 & 86.7 &-
& 12.1 & 83.3 & -& 8.0 & 93.9 &-
& \textcolor{black}{24.6} & \textcolor{black}{\cellcolor{best}78.0} & -
& \textcolor{black}{5.3}               \\

GenPercept$^{\star^\S}$       
 &\cellcolor{best2}74K
& 5.6 & 96.0 & 99.2 & 13.0 & 84.2 & 97.2
& 7.0 & 95.6 & \cellcolor{best2}98.8 & 6.2 & 96.1 & 99.1
& \textcolor{black}{35.7} & \textcolor{black}{75.6} & \textcolor{black}{86.6}
& \textcolor{black}{4.9}             \\

\textcolor{black}{Diffusion-E2E-FT$^{\star}$} & \textcolor{black}{\cellcolor{best2}74K} & \textcolor{black}{5.4} & \textcolor{black}{96.5} & \textcolor{black}{99.1} & \textcolor{black}{9.6} & \textcolor{black}{92.1} & \textcolor{black}{98.0} & \textcolor{black}{\cellcolor{best2}6.4} & \textcolor{black}{\cellcolor{best2}95.9} & \textcolor{black}{98.7} & \textcolor{black}{5.8} & \textcolor{black}{96.5} & \textcolor{black}{98.8} & \textcolor{black}{30.3} & \textcolor{black}{\cellcolor{best2}77.6} & \textcolor{black}{\cellcolor{best}87.9} & \textcolor{black}{3.6} \\

\textcolor{black}{DepthAnything V2 }  
& \textcolor{black}{62.6M}
& \textcolor{black}{\cellcolor{best2}4.5} & \textcolor{black}{\cellcolor{best2}97.9} & \textcolor{black}{\cellcolor{best2}99.3}
& \textcolor{black}{\cellcolor{best}7.4} & \textcolor{black}{\cellcolor{best2}94.6} & \textcolor{black}{98.6}
& \textcolor{black}{13.1} & \textcolor{black}{86.5} & \textcolor{black}{97.5}
& \textcolor{black}{\cellcolor{best}4.2} & \textcolor{black}{\cellcolor{best2}97.8} & \textcolor{black}{\cellcolor{best2}99.3}
& \textcolor{black}{26.5} & \textcolor{black}{73.4} & \textcolor{black}{87.1}
& \textcolor{black}{3.5} \\

\textcolor{black}{\textbf{Lotus-D (Ours)}$^{\star}$} 
& \textcolor{black}{\cellcolor{best}\textbf{59K}}
& \textcolor{black}{\textbf{5.1}} & \textcolor{black}{\textbf{97.2}} & \textcolor{black}{\textbf{99.2}} 
& \textcolor{black}{\textbf{8.1}} & \textcolor{black}{\textbf{93.1}} & \textcolor{black}{\cellcolor{best2}\textbf{98.7}} 
& \textcolor{black}{\cellcolor{best}\textbf{6.1}} & \textcolor{black}{\cellcolor{best}\textbf{97.0}} & \textcolor{black}{\cellcolor{best}\textbf{99.1}} 
& \textcolor{black}{\textbf{5.5}} & \textcolor{black}{\textbf{96.5}} & \textcolor{black}{\textbf{99.0}} 
& \textcolor{black}{\cellcolor{best2}\textbf{22.8}} & \textcolor{black}{\textbf{73.8}} & \textcolor{black}{\textbf{86.2}} 
& \textcolor{black}{\cellcolor{best2}\textbf{3.0}}  \\

\textcolor{black}{DepthAnything}  
& \textcolor{black}{62.6M}
& \textcolor{black}{\cellcolor{best}4.3} & \textcolor{black}{\cellcolor{best}98.1} & \textcolor{black}{\cellcolor{best}99.6}
& \textcolor{black}{\cellcolor{best2}7.6} & \textcolor{black}{\cellcolor{best}94.7} & \textcolor{black}{\cellcolor{best}99.2}
& \textcolor{black}{12.7} & \textcolor{black}{88.2} & \textcolor{black}{98.3}
& \textcolor{black}{\cellcolor{best2}4.3} & \textcolor{black}{\cellcolor{best}98.1} & \textcolor{black}{\cellcolor{best}99.6}
& \textcolor{black}{26.0} & \textcolor{black}{75.9} & \textcolor{black}{\cellcolor{best2}87.5}
& \textcolor{black}{\cellcolor{best}2.4}  \\
\midrule

GeoWizard$^{\star^\S}$
 &280K
& 5.6 & 96.3 & \cellcolor{best2}99.1 & 14.4 & 82.0 & 96.6
& 6.6 & 95.8 & 98.4 & \cellcolor{best2}6.4 & 95.0 & 98.4
& \textcolor{black}{33.5} & \textcolor{black}{72.3} & \textcolor{black}{86.5}
&  \textcolor{black}{3.3}                \\

Marigold$_\text{(LCM)}$$^{\star\S}$
 &74K
&6.1 & 95.8 & 99.0 & \cellcolor{best2}9.8 & \cellcolor{best2}91.8 & \cellcolor{best}98.7
& 6.8 & 95.6 & \cellcolor{best2}99.0& 6.9& 94.6 &98.6
& \cellcolor{best2}\textcolor{black}{30.7} & \textcolor{black}{\cellcolor{best}77.5} & \textcolor{black}{\cellcolor{best}89.3}
&  \textcolor{black}{2.9}              \\

Marigold$^{\star}$
&74K
& \cellcolor{best2}5.5 & \cellcolor{best2}96.4 & \cellcolor{best2}99.1 & 9.9 & 91.6 &\cellcolor{best} 98.7
& \cellcolor{best2}6.5 & \cellcolor{best2}95.9 & \cellcolor{best2}99.0 & \cellcolor{best2}6.4 & \cellcolor{best2}95.2 & \cellcolor{best}98.8
& \textcolor{black}{30.8} & \cellcolor{best2}\textcolor{black}{77.3} & \textcolor{black}{\cellcolor{best2}88.7}
&  \cellcolor{best2} \textcolor{black}{2.1 }              \\

\textcolor{black}{\textbf{Lotus-G (Ours)}$^{\star}$} & \textcolor{black}{\textbf{59K}} & \textcolor{black}{\cellcolor{best}\textbf{5.4}} & \textcolor{black}{\cellcolor{best}\textbf{96.8}} & \textcolor{black}{\cellcolor{best}\textbf{99.2}} 
& \textcolor{black}{\cellcolor{best}\textbf{8.5}} & \textcolor{black}{\cellcolor{best}\textbf{92.2}} & \textcolor{black}{\cellcolor{best2}\textbf{98.4}} 
& \textcolor{black}{\cellcolor{best}\textbf{5.9}} & \textcolor{black}{\cellcolor{best}\textbf{97.0}} & \textcolor{black}{\cellcolor{best}\textbf{99.2}} 
& \textcolor{black}{\cellcolor{best}\textbf{5.9}} & \textcolor{black}{\cellcolor{best}\textbf{95.7}} & \textcolor{black}{\cellcolor{best}\textbf{98.8}} 
& \textcolor{black}{\cellcolor{best}\textbf{22.9}} & \textcolor{black}{\textbf{72.9}} & \textcolor{black}{\textbf{86.0}} & \textcolor{black}{\cellcolor{best}\textbf{1.3}} \\

\bottomrule
\end{tabular}
}
\end{table}

\setlength{\tabcolsep}{2pt}
\begin{table}[h]
\scriptsize
\caption{\textbf{Quantitative comparison on zero-shot surface normal estimation} between Lotus and SoTA methods. Discriminative methods are shown in the upper section, generative methods in the lower. Both \textbf{Lotus-D} and \textbf{Lotus-G} outperform all other methods. $^\ddag$refers the Marigold normal model as detailed in this \href{https://huggingface.co/prs-eth/marigold-normals-lcm-v0-1}{link}. $^\star$denotes the method relies on pre-trained Stable Diffusion. }
\vspace{-3mm}
\label{tab:normal}
\scriptsize
\centering
\resizebox{\textwidth}{!}{
\begin{tabular}{l|c|ccc|ccc|ccc|ccc|ccc|c}
\toprule
\multirow{2}{*}{Method} & Training
& \multicolumn{3}{c|}{NYUv2 (Indoor)} & \multicolumn{3}{c|}{ScanNet (Indoor)} 
& \multicolumn{3}{c|}{iBims-1 (Indoor)} & \multicolumn{3}{c|}{Sintel (Outdoor)}
& \multicolumn{3}{c|}{\textcolor{black}{OASIS (Various)}}
& \textcolor{black}{Avg.} \\
 &Data$\downarrow$
 & m.$\downarrow$ & $11.25^\circ$$\uparrow$ & $30^\circ$$\uparrow$ 
 & m.$\downarrow$ & $11.25^\circ$$\uparrow$ & $30^\circ$$\uparrow$
 & m.$\downarrow$ & $11.25^\circ$$\uparrow$ & $30^\circ$$\uparrow$
 & m.$\downarrow$ & $11.25^\circ$$\uparrow$ & $30^\circ$$\uparrow$
&\textcolor{black}{ m.$\downarrow$} & \textcolor{black}{$11.25^\circ$$\uparrow$} & \textcolor{black}{$30^\circ$$\uparrow$}
 &    \textcolor{black}{Rank}           \\
\midrule
OASIS
&110K
& 29.2 & 23.8 & 60.7
& 32.8 & 15.4 & 52.6
& 32.6 & 23.5 & 57.4
& 43.1 & 7.0  & 35.7
& - & - & -
&  \textcolor{black}{7.8}                 \\

Omnidata
 &12.2M
& 23.1 & 45.8 & 73.6
& 22.9 & 47.4 & 73.2
& 19.0 & 62.1 & 80.1
& 41.5 & 11.4 & 42.0
& \textcolor{black}{24.9} & \cellcolor{best2}\textcolor{black}{31.0} & \textcolor{black}{71.4}
&  \textcolor{black}{5.9}                 \\

EESNU
&2.5M
& \cellcolor{best}16.2 & 58.6 & 83.5
& -    & -    & -
& 20.0 & 58.5 & 78.2
& 42.1 & 11.5 & 41.2
& \textcolor{black}{27.7} & \textcolor{black}{24.0} & \textcolor{black}{66.6}
&  \textcolor{black}{5.8}                 \\

GenPercept$^{\S\star}$
&74K
& 18.2 & 56.3 & 81.4
& 17.7 & 58.3 & 82.7
& \cellcolor{white}18.2 & \cellcolor{white}64.0 & \cellcolor{white}82.0
& \cellcolor{white}37.6 & \cellcolor{white}16.2 & \cellcolor{white}51.0
& \textcolor{black}{26.3} & \textcolor{black}{26.9} & \textcolor{black}{71.1}
& \textcolor{black}{4.9}                \\

Omnidata V2
 &12.2M
& 17.2 & 55.5 & 83.0
& 16.2 & 60.2 & 84.7
& 18.2 & 63.9 & 81.1
& 40.5 & 14.7 & 43.5
& 24.2 & 27.7 & 74.2
& 4.4                \\

\cellcolor{white}DSINE
&160K
& 16.4 & 59.6 & \cellcolor{best2}83.5 
& 16.2 & 61.0 & \cellcolor{white}84.4
& \cellcolor{best2}17.1 & \cellcolor{best2}67.4 & 82.3
&34.9 & 21.5 & 52.7
&\textcolor{black}{24.4} & \textcolor{black}{28.8} & \textcolor{black}{72.0}
& \textcolor{black}{3.1}                \\

\rowcolor{white}
\cellcolor{white}\textcolor{black}{Diffusion-E2E-FT$^{\S\star}$} 
& \cellcolor{best2}\textcolor{black}{74K}
& \textcolor{black}{16.5} & \cellcolor{best}\textcolor{black}{60.4} & \textcolor{black}{83.1}
& \cellcolor{best}\textcolor{black}{14.7} & \cellcolor{best}\textcolor{black}{66.1} & \cellcolor{best2}\textcolor{black}{85.1}
& \cellcolor{best}\textcolor{black}{16.1} & \cellcolor{best}\textcolor{black}{69.7} & \cellcolor{best}\textcolor{black}{83.9}
&\cellcolor{best2} \textcolor{black}{33.5} & \cellcolor{best2}\textcolor{black}{22.3} & \cellcolor{best2}\textcolor{black}{53.5}
& \cellcolor{best2}\textcolor{black}{23.2} & \textcolor{black}{29.4} & \cellcolor{best2}\textcolor{black}{74.5}
& \cellcolor{best2}\textcolor{black}{1.9}   \\

\rowcolor{best}
\cellcolor{white}\textbf{Lotus-D (Ours)}$^\star$ 
&\cellcolor{best}\textbf{59K}
& \cellcolor{best}\textbf{16.2} & \cellcolor{best2}\textbf{59.8} & \textbf{83.9} & \cellcolor{best}\textbf{14.7}
& \cellcolor{best2}\textbf{64.0} & \textbf{86.1} & \cellcolor{best2}\textbf{17.1} & \cellcolor{white}\textbf{66.4}
& \cellcolor{best2}\textbf{83.0} &\cellcolor{best} \textbf{32.3} & \cellcolor{best}\textbf{22.4} & \textbf{57.0}
& \cellcolor{best}\textcolor{black}{\textbf{22.3}} & \cellcolor{best}\textcolor{black}{\textbf{31.8}} & \cellcolor{best}\textcolor{black}{\textbf{76.1}}
& \cellcolor{best}\textcolor{black}{\textbf{1.4}}  \\

\midrule
\cellcolor{white}Marigold$^{\ddag\star}$
&\cellcolor{best2}74K
& 20.9 & 50.5 &
-& 21.3 & 45.6 &
-& \cellcolor{white}18.5 & \cellcolor{white}64.7 &
-& -  & -    & -
& - & - & -
& \textcolor{black}{3.6}
\\

\rowcolor{white}
\cellcolor{white}GeoWizard$^{\S\star}$
&\cellcolor{white}280K
& 18.9 & 50.7 & 81.5
& 17.4 & 53.8 & 83.5
& \cellcolor{white}19.3 & \cellcolor{white}63.0 & \cellcolor{white}80.3
& 40.3 & 12.3 & 43.5
& \cellcolor{best2}\textcolor{black}{25.2} & \textcolor{black}{23.4} & \textcolor{black}{68.1}
& \textcolor{black}{3.1}                \\

\rowcolor{best2}
\cellcolor{white}StableNormal$^{\S\star}$
&\cellcolor{white}250K
& 18.6 & 53.5 & 81.7
& 17.1 & 57.4 & 84.1
& 18.2 & 65.0 & 82.4
& 36.7 & 14.1 & 50.7
& \cellcolor{white}\textcolor{black}{26.5} & \textcolor{black}{23.5} & \textcolor{black}{68.7}
& \textcolor{black}{2.1}  \\

\rowcolor{best}
\cellcolor{white}\textbf{Lotus-G (Ours)}$^\ast$ 
&\textbf{59K}
&  \textbf{16.5}&  \textbf{59.4}&\textbf{83.5}
&  \textbf{15.1}&  \textbf{63.9}&\textbf{85.3}
&  \textbf{17.2}&  \textbf{66.2}&\textbf{82.7}
&  \textbf{33.6}&  \textbf{21.0}&\textbf{53.8} 
& \textcolor{black}{\textbf{22.7}} & \textcolor{black}{\textbf{29.4}} & \textcolor{black}{\textbf{75.8}}
& \textcolor{black}{\textbf{1.0}}   \\

\bottomrule
\end{tabular}
}
\end{table}

\setlength{\tabcolsep}{1.pt}
\begin{table}[]
\scriptsize
\caption{\textbf{Ablation studies} on the step-by-step design of our adaptation protocol for fitting pre-trained diffusion models into dense prediction. Here we show the results in monocular depth estimation.}
\vspace{-3mm}
\label{tab:ablation}
\begin{tabular}{l|c|ccc|ccc|ccc|ccc}
\toprule
\multirow{2}{*}{Method} & Training
& \multicolumn{3}{c|}{NYUv2 (Indoor)} & \multicolumn{3}{c|}{KITTI (Outdoor)} & \multicolumn{3}{c|}{ETH3D (Various)} & \multicolumn{3}{c}{ScanNet (Indoor)} \\
& Data
& AbsRel$\downarrow$ & $\delta$1$\uparrow$ & $\delta$2$\uparrow$ 
& AbsRel$\downarrow$ & $\delta$1$\uparrow$ & $\delta$2$\uparrow$ 
& AbsRel$\downarrow$ & $\delta$1$\uparrow$ & $\delta$2$\uparrow$ 
& AbsRel$\downarrow$ & $\delta$1$\uparrow$ & $\delta$2$\uparrow$ 
\\
\midrule
Direct Adaptation   & 39K          
& 11.551 & 87.692 & 96.122 
& 20.164 & 70.403 & 90.996
& 19.894 & 76.464 & 87.960
& 15.726 & 78.885 & 93.651 \\
$+$ $x_0$-prediction    & 39K
& 8.332  & 92.769 & 97.941  
& 17.008 & 74.969 & 93.611
& 11.075 & 87.952 & 94.978
& 10.212& 89.130 & 97.181 \\
$+$ Single Time-step     
& 39K
& 5.587  & 96.272 & 99.113   
& 13.262 & 83.210 & 97.237
& 7.586 & 94.143 & 97.678
& 6.262 & 95.394 & 98.791 \\
$+$ Detail Preserver     & 39K       
& 5.555 & 96.303 & 99.118
& 13.170  & 83.657 & 97.454
& 7.147 & 95.000 & 98.058
& 6.201 & 95.470 & 98.814 \\
\midrule
$+$ Mixture Dataset  & 59K
& 5.425 & 96.597 & 99.156
& 11.324 & 87.692 & 97.780
& 6.172 & 96.077 & 98.980
& 6.024 & 96.026 & 99.730 \\
\tiny \hspace{2mm}$\hookrightarrow$ $-$ Noise Input  & 59K  
& 5.334 & 96.729 & 99.198
& 9.334 & 92.813 & 98.795
&  6.846 & 95.290 & 98.899
&  5.982 & 96.287 & 99.087 \\
\rowcolor{best2}\cellcolor{white}\textcolor{black}{$+$ Disparity Space \textbf{(Lotus-G})}   & \cellcolor{white}\textcolor{black}{59K}
& \textcolor{black}{5.379} & \textcolor{black}{96.736} & \cellcolor{best}\textcolor{black}{99.155}
& \textcolor{black}{8.521} & \textcolor{black}{92.206} & \textcolor{black}{98.374}
& \textcolor{black}{5.878} & \textcolor{black}{97.024} & \textcolor{black}{99.233}
& \textcolor{black}{5.925} & \textcolor{black}{95.727} & \textcolor{black}{98.839} \\
\rowcolor{best}\cellcolor{white}
\tiny \textcolor{black}{\hspace{2mm}$\hookrightarrow$ $-$ Noise Input (\textbf{Lotus-D})}     & \cellcolor{white}\textcolor{black}{59K}  
& \textcolor{black}{5.123} & \textcolor{black}{97.182} & \cellcolor{best2}\textcolor{black}{99.134}
& \textcolor{black}{8.117} & \textcolor{black}{93.097} & \textcolor{black}{98.654}
& \textcolor{black}{6.147} & \textcolor{black}{96.964} & \textcolor{black}{99.077}
& \textcolor{black}{5.494} & \textcolor{black}{96.534} & \textcolor{black}{99.039} \\
\bottomrule
\end{tabular}
\end{table}

\section{Conclusion and Future Work}
In this paper, we introduce Lotus, a diffusion-based visual foundation model for dense prediction. \haodong{Through systematic analysis and tailored diffusion formulation, Lotus finds a way to better fit the rich visual prior from pre-trained diffusion models into dense prediction.}
Extensive experiments demonstrate that Lotus achieves promising performance on zero-shot depth and normal estimation with minimal training data, paving the way of various practical applications. Please see the supplementary materials for our discussion about \textbf{Applications} (Sec.~\ref{suppl:appl}) and \textbf{Future Work} (Sec.~\ref{suppl:future}).

{\fontsize{16pt}{20pt}\selectfont\sc {Supplementary Materials of \\ \textbf{Lotus}: Diffusion-based Visual Foundation Model for High-quality Dense Prediction}\par}

\appendix

\section{Experimental Settings}
\subsection{Implementation Details}
\label{suppl:impl}
We implement Lotus based on Stable Diffusion V2~\citep{rombach2022high}, with text conditioning disabled. Both the depth and normal maps are normalized to the range $[-1,1]$ to match the designed input value range of the VAE. During training, we fix the time-step $t=1000$. To optimize the model, we utilize the standard Adam optimizer with the learning rate $3\times 10^{-5}$. All experiments are conducted on 8 NVIDIA A800 GPUs and the total batch size is 128. For our discriminative variant, we train for 4,000 steps, which takes $\sim$8.1 hours, while for the generative variant, we extend training to 10,000 steps, requiring $\sim$20.3 hours.

\subsection{Evaluation Datasets and Metrics}
\label{suppl:eval_metrics}
\textbf{Evaluation Datasets.}
\ding{172} For affine-invariant depth estimation, we evaluate on 4 real-world datasets that are not seen during training: NYUv2~\citep{silberman2012indoor} and ScanNet~\citep{dai2017scannet} all contain images of indoor scenes; KITTI~\citep{geiger2013vision} contains various outdoor scenes; ETH3D~\citep{schops2017multi}, a high-resolution dataset, containing both indoor and outdoor scenes.
\ding{173} For surface normal prediction, we employ 4 datasets for evaluation: NYUv2~\citep{silberman2012indoor}, ScanNet~\citep{dai2017scannet}, and iBims-1~\citep{koch2018evaluation} contain real indoor scenes; Sintel~\citep{butler2012naturalistic} contains highly dynamic outdoor scenes.

\textbf{Metrics.}
\ding{172} For affine-invariant depth, we follow the evaluation protocol from~\citep{ranftl2020towards,ke2024repurposing,yang2024depth1,yang2024depth2}, aligning the estimated depth predictions with available ground truths using least-squares fitting. The accuracy of the aligned predictions is assessed using the \emph{absolute mean relative error} (AbsRel), \emph{i.e.}, $\frac{1}{M}\sum_{i=1}^M |a_i-d_i|/d_i$, where $M$ is the total number of pixels, $a_i$ is the predicted depth map and $d_i$ represents the ground truth. We also report $\delta 1$ and $\delta2$, the proportion of pixels satisfying $\text{Max}(a_i/d_i, d_i/a_i)<1.25$ and $< 1.25^2$ respectively. 

\ding{173} For surface normal, following~\citep{bae2024dsine,ye2024stablenormal}, we evaluate the predictions of Lotus by measuring the mean angular error for pixels with available ground truth. Additionally, we report the percentage of pixels with an angular error below $11.25^{\circ}$ and $30^\circ$. 

For all tasks, we report the \emph{Avg. Rank}, which indicates the average ranking of each method across various datasets and evaluation metrics. A lower value signifies better overall performance.

\section{Details of Direct Adaption}
\label{suppl:da}
As illustrated in Fig.~\ref{fig:dada} of the main paper, our Direct Adaption means directly adapting the standard diffusion formulation for dense prediction task with minimal modifications. 
Specifically, starting with the pre-trained Stable Diffusion model, image $\textbf{x}$ and annotation $\textbf{y}$ are encoded using the pre-trained VAE encoder. Noise is added to the encoded annotation to obtain the noisy annotation $\mathbf{z_t^y}$ at noise level $t \in [1,T]$. The encoded image $\mathbf{z^x}$ is then concatenated with the noisy annotation $\mathbf{z_t^y}$ to form the input of the denoiser U-Net model. To handle this concatenated input, the U-Net input layer is duplicated (from 4 channels to 8 channels) and its original weights are halved as initialization, which prevents activation inflation~\citep{ke2024repurposing}. Direct Adaptation is optimized using the standard multi-step formulation the standard diffusion objective, $\epsilon$-prediction, as described in Eq.~\ref{eq:eps} (first row) of the main paper. To analyze the original diffusion formulation more effectively, we avoid specialized techniques introduced in prior methods \citep{ke2024repurposing, fu2024geowizard,xu2024diffusion,ye2024stablenormal}, such as annealed multi-resolution noise (AMRN). 

\begin{table}[t]
\centering
\caption{Experiments based on Marigold $w/$ AMRN. }
\label{tab:w_amrn}
\setlength{\tabcolsep}{6pt}
\begin{tabular}{l|l|cc|cc}
\toprule
\multirow{2}{*}{Index} & \multirow{2}{*}{Method} & \multicolumn{2}{c|}{NYUv2} & \multicolumn{2}{c}{KITTI} \\
                       &                        & AbsRel$\downarrow$ & $\delta1$$\uparrow$ & AbsRel$\downarrow$ & $\delta1$$\uparrow$ \\
\midrule
1-1 & $\epsilon$-pred. & 6.746 & 95.021 & 11.827 & 87.065 \\
1-2 & $\epsilon$-pred. + single step & 6.691 & 94.552 & 13.395 & 76.269 \\
1-3 & $\epsilon$-pred. + single step + detail preserver & 6.547 & 94.772 & 12.815 & 77.829 \\
\midrule
2-1 & $v$-pred. & 6.358 & 95.188 & 10.796 & 89.726 \\
2-2 & $v$-pred. + single step & 5.499 & 96.415 & 11.132 & 88.520 \\
2-3 & $v$-pred. + single step + detail preserver & 5.422 & 96.517 & 10.761 & 89.826 \\
\midrule
3-1 & $x_0$-pred. & 6.262 & 95.501 & 10.769 & 89.643 \\
3-2 & $x_0$-pred. + single step & 5.495 & 96.431 & 11.237 & 88.457 \\
\textbf{3-3} & \textbf{$x_0$-pred. + single step + detail preserver} & \textbf{5.418} & \textbf{96.542} & \textbf{10.651} & \textbf{89.887} \\
\bottomrule
\end{tabular}
\end{table}

\begin{table}[t]
\centering
\caption{Experiments based on Marigold $w/o$ AMRN. }
\label{tab:wo_amrn}
\setlength{\tabcolsep}{6pt}
\begin{tabular}{l|l|cc|cc}
\toprule
\multirow{2}{*}{Index} & \multirow{2}{*}{Method} & \multicolumn{2}{c|}{NYUv2} & \multicolumn{2}{c}{KITTI} \\
                       &                        & AbsRel$\downarrow$ & $\delta1$$\uparrow$ & AbsRel$\downarrow$ & $\delta1$$\uparrow$ \\
\midrule
1-1 & $\epsilon$-pred. & 13.110 & 85.083 & 17.655 & 75.581 \\
1-2 & $\epsilon$-pred. + single step & 6.605 & 94.583 & 13.406 & 76.298 \\
1-3 & $\epsilon$-pred. + single step + detail preserver & 6.582 & 94.768 & 12.823 & 77.983 \\
\midrule
2-1 & $v$-pred. & 10.634 & 89.448 & 14.328 & 84.026 \\
2-2 & $v$-pred. + single step & 5.498 & 96.562 & 11.173 & 88.314 \\
2-3 & $v$-pred. + single step + detail preserver & 5.459 & 96.657 & 10.814 & 89.081 \\
\midrule
3-1 & $x_0$-pred. & 8.058 & 92.834 & 12.177 & 86.301 \\
3-2 & $x_0$-pred. + single step & 5.477 & 96.615 & 11.166 & 88.640 \\
\textbf{3-3} & \textbf{$x_0$-pred. + single step + detail preserver} & \textbf{5.396} & \textbf{96.717} & \textbf{10.575} & \textbf{89.804} \\
\bottomrule
\end{tabular}
\end{table}

The AMRN strategy aims to reduce the model's variance, which has a similar effect to our design, $x_0$-pred., but through a different solution. This diminishes the impact of our method. Therefore, it is preferable to validate the effect of our designs $w/o$ AMRN. We validate this claim using the Marigold codebase, both $w/$ and $w/o$ AMRN, as shown in the Tab.~\ref{tab:w_amrn} and Tab.~\ref{tab:wo_amrn}, respectively. 
In Tab.~\ref{tab:wo_amrn}, the performance of multi-step models follows the order: $\epsilon$-pred. $<$ $v$-pred. $<$ $x_0$-pred. However, in Tab.~\ref{tab:w_amrn}, the differences between three parameterization types are minimal, particularly the performance of $v$-pred. and $x_0$-pred. are nearly identical. This can be attributed to the influence of AMRN, which is specifically designed for multi-step diffusion models to reduce variance and enhance performance. As a result, $x_0$-pred. shows no significant difference in reducing variance compared to the other two parameterizations. In Tab.~\ref{tab:wo_amrn}, when the number of time-steps is reduced to one, the performance of the model improves regardless of the parameterization type used. However, in Tab.~\ref{tab:w_amrn}, the effect of single-step is unstable. This unexpected phenomenon arises from the complex, multifaceted effects of AMRN when transitioning from multi-step to single-step: \ding{172} AMRN significantly improves the multi-step model, but its effect is lost when the number of time-steps is reduced to one. \ding{173} In the single-step model, convergence is easier with limited data, leading to a slight improvement in performance. However, this also leads to catastrophic forgetting, which reduces the model's ability to handle detailed areas, especially on the KITTI dataset. In both Tab.~\ref{tab:w_amrn} and Tab.~\ref{tab:wo_amrn}, Detail Preserver further enhances the performance of single-step model, particularly on the KITTI dataset, which contains more complex and detailed areas, such as pedestrians and fences, compared to the NYUv2 dataset. In both Tab.~\ref{tab:w_amrn} and Tab.~\ref{tab:wo_amrn}, when using a single step ($t=T$), according to $\mathbf{v}_t = \sqrt{\bar{\alpha}_T} \boldsymbol{\epsilon} - \sqrt{1 - \bar{\alpha}_T} \mathbf{z}$, since $\sqrt{\bar{\alpha}_T} \approx 0$ when $t = T$, $v$-pred. becomes equivalent to $x_0$-pred. This explains why the performances of $v$-pred. and $x_0$-pred. are nearly identical in single-step, with only minor differences. In conclusion, these experiments show that AMRN, which has a similar effect to our designs but is achieved through a different solution, diminishing the impact of our proposed designs. Therefore, it is preferable to validate the effect of our designs $w/o$ AMRN. The experiments on Marigold $w/o$ AMRN (Tab.~\ref{tab:wo_amrn}) validate the effectiveness of our proposed designs, as stated in our main paper, where the best protocol is $x_0$-pred. + single step + detail preserver. 

\section{Analysis of ``$\text{direction}(\mathbf{z^y_{\tau}})$'' in DDIM Process (Eq.~\ref{eq:param_types})}
\label{suppl:direction}
In addition to the predicted clean sample $\mathbf{\hat{z}^y_\tau}$, Eq.~\ref{eq:param_types} of the main paper includes another term, ``direction($\mathbf{z^y_{\tau}}$)''. It is calculated according to different parameterization types:
\begin{equation}
\label{eq:direction}
\begin{aligned}
\epsilon\text{-prediction: } d &= w_\tau \cdot f_\theta^\epsilon \\
x_0\text{-prediction: } d &= w_\tau \cdot [ \frac{1}{\sqrt{1-\overline{\alpha}_\tau}}(\mathbf{z^y_\tau}-\sqrt{\overline{\alpha}_\tau} f_\theta^\textbf{z})]
\end{aligned}
\end{equation}
where $d$ represents the term ``direction($\mathbf{z^y_{\tau}}$)'', $w_\tau=\sqrt{1-\overline{\alpha}_{\tau-1}}$ is the weight at denoising step $\tau$. And $f_\theta^\epsilon$ and $f_\theta^\textbf{z}$ denote the model outputs for different parameterizations. For clarity, the input of the model $f_\theta$ is omitted. 
As shown in Eq.~\ref{eq:direction}, for $x_0$-prediction, when $\tau\rightarrow 1$, \emph{i.e.}, at the end of the denoising process, the factor $\sqrt{1-\overline{\alpha}_\tau} \rightarrow 0$, which may amplify variance from $f_\theta^\textbf{z}$. However, its influence is limited. The reasons are as follows:
\ding{172} The rate of change of $\sqrt{1-\overline{\alpha}_\tau}$ from $T$ to 1 is initially slow and then accelerates. As a result, the factor remains close to $1$ for most of the denoising process, only close to $0$ in the final steps.
\ding{173} In $x_0$-prediction, compared to the initial denoising steps, the gap between network output $f^{\mathbf{z}}_\theta$ and $\mathbf{z^y_\tau}$ in the final steps is much weaker and gradually approaching zero. With $\sqrt{\overline{\alpha}_\tau}\rightarrow 1$ as $\tau\rightarrow1$, we can get $\mathbf{z^y_\tau}-\sqrt{\overline{\alpha}_\tau} f_\theta^\textbf{z}\rightarrow0$, which may also indicate the limited influence of factor $\sqrt{1-\overline{\alpha}_\tau}$.

\section{Performance of $v$-prediction}
\label{suppl:v-pred}
\begin{wrapfigure}{t}{0.5\textwidth}
    \centering
    \vspace{-5mm}
    \includegraphics[width = \linewidth]{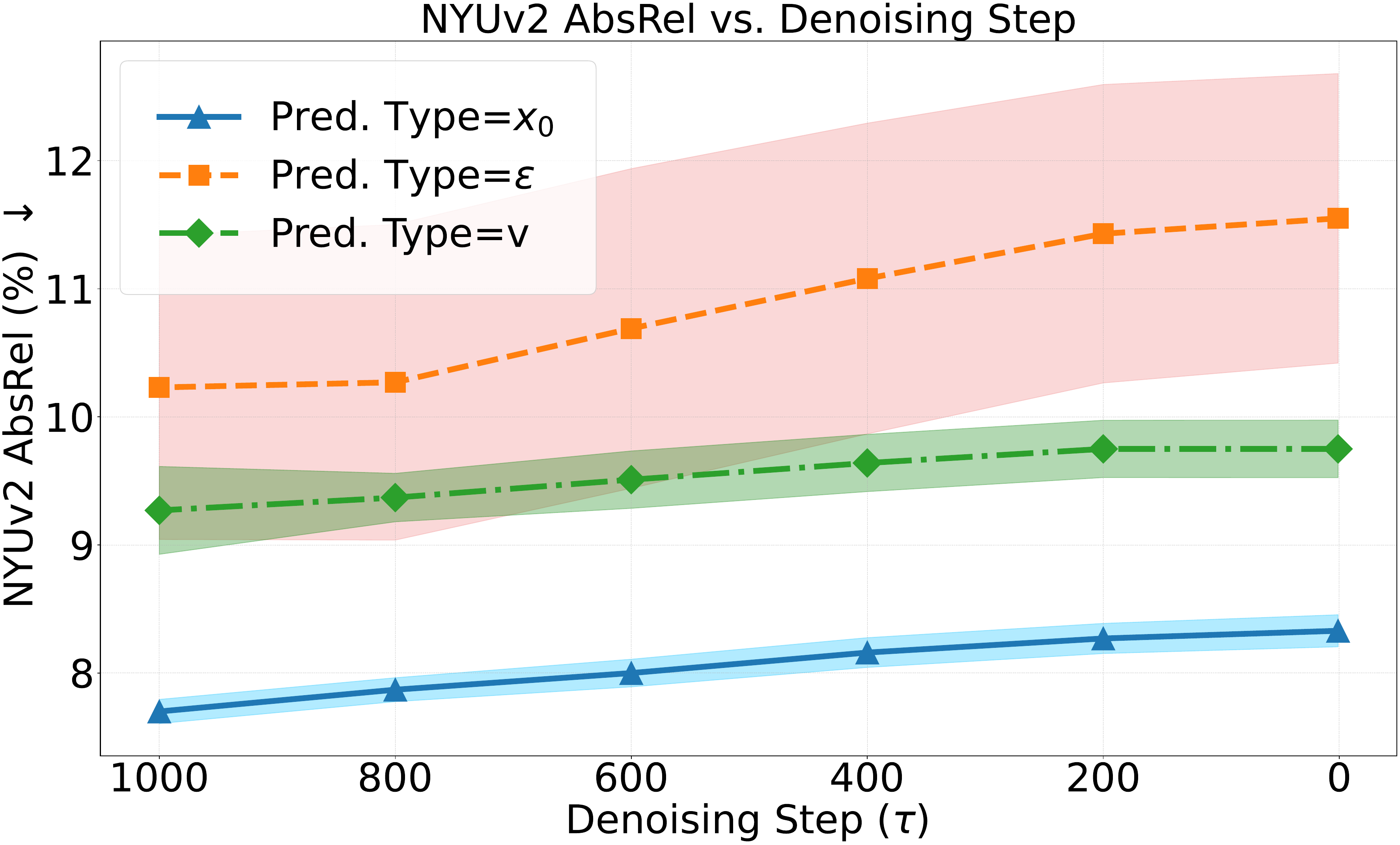}
    \caption{\textbf{Quantitative evaluation of the predicted depth maps $\mathbf{\hat{z}^y_\tau}$ along the denoising process.} The experimental settings are same as Fig.~\ref{fig:3param} and~\ref{fig:pixe_error_with_variance}. Six steps are selected for illustration. The banded regions around each line indicate the variance, wider areas representing larger variance. 
    }
    \vspace{-5mm}
\label{fig:pixe_error_v_var}
\end{wrapfigure}
In sec.~\ref{sec:param}, we discussed two basic parameterization types: $\epsilon$-prediction and $x_0$-prediction. The latest parameterization, $v$-prediction~\citep{salimans2022progressive}, combines these two basic parameterizations to avoid the invalid prediction values of $\epsilon$-prediction at some time-steps for progressive distillation.
Specifically, the U-Net denoiser model $f_{\theta}$ learns to predict the combination of added noise $\epsilon$ and the clean sample $\mathbf{z^y}$: $\textbf{v} = \sqrt{\overline{\alpha}_\tau}\epsilon-\sqrt{1-\overline{\alpha}_\tau} \mathbf{z^y}$, where ${\sqrt{\overline{\alpha}_\tau}}^2+{\sqrt{1-\overline{\alpha}_\tau}}^2=1$. 
During inference, according to the Eq.~\ref{eq:param_types} of main paper, the prediction $\mathbf{\hat{z}^y_{\tau}} = \sqrt{\overline{\alpha}_\tau}\mathbf{z^y_\tau}- \sqrt{1-\overline{\alpha}_\tau}f_\theta^\textbf{v}$, where $f_\theta^\textbf{v}$ represents the predicted combination, striking a balance between $\epsilon$ ($\epsilon$-prediction) and $\mathbf{z^y}$ ($x_0$-prediction).
As shown in Fig.~\ref{fig:pixe_error_v_var}, we conduct experiments based on the settings in Fig.~\ref{fig:3param} and~\ref{fig:pixe_error_with_variance} of the main paper. The results indicate that the performance of $v$-prediction falls between that of $x_0$-prediction and $\epsilon$-prediction, with moderate variance.
However, for dense prediction tasks, minimizing variance is crucial to avoid unstable prediction. Therefore, $v$-prediction may not be the optimal choice. In contrast, $x_0$-prediction achieves the best performance with the lowest variance, which is why we replace the standard $\epsilon$-prediction with the more suitable $x_0$-prediction.

\section{Experiments on More Dense Prediction Tasks:\\$\quad\ \ $Semantic Segmentation and Diffuse Reflectance}
\begin{figure}[!ht]
    \centering
    \includegraphics[width = \linewidth]{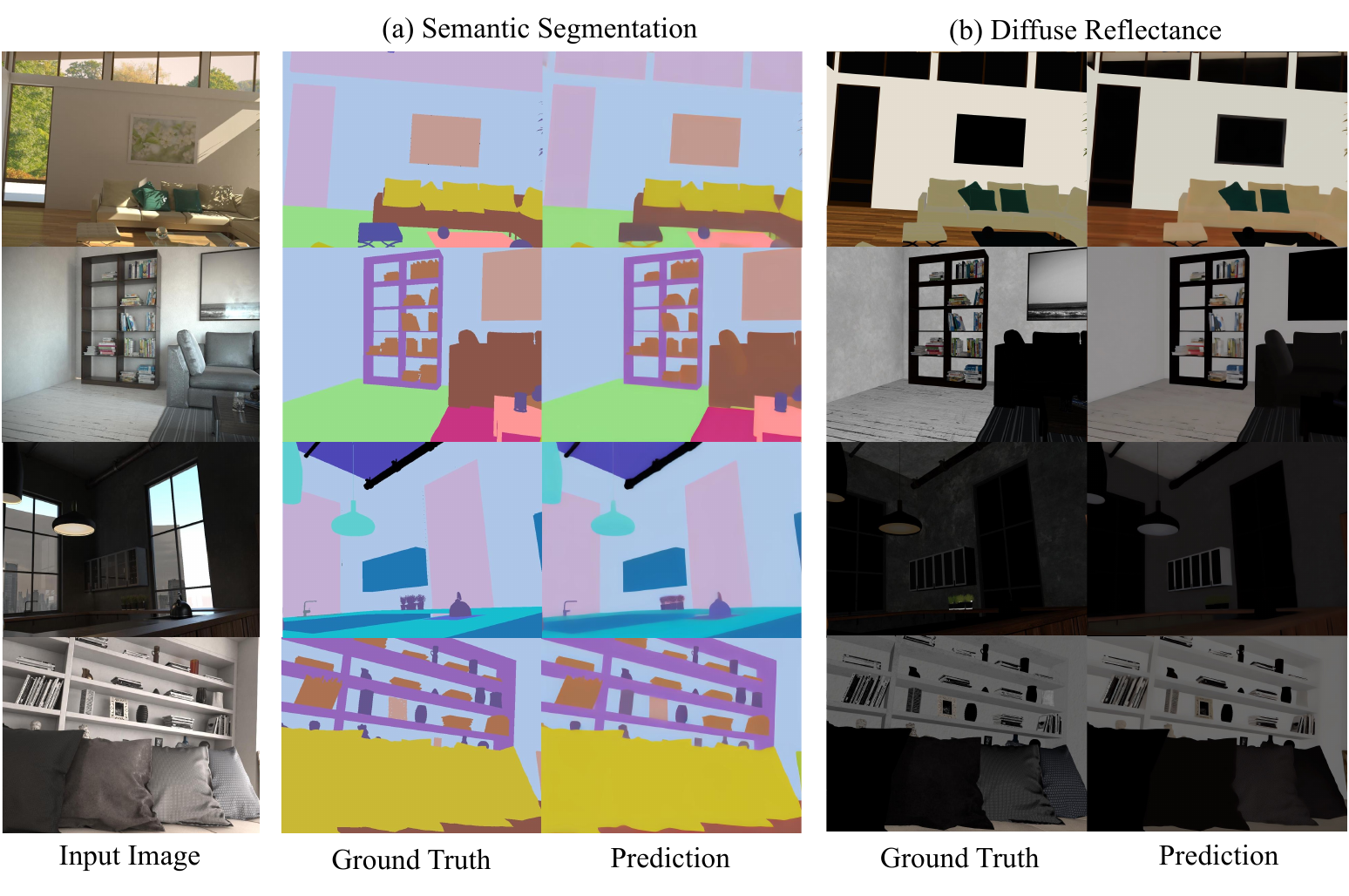}
    \caption{\textbf{Experiments of Lotus on (a) semantic segmentation and (b) diffuse reflectance.} The high-quality results indicate that our method, even without task-specific designs, can be effectively applied not only to geometric dense prediction tasks, but also to semantic dense prediction tasks.  }
    \label{fig:other_tasks}
\end{figure}

\begin{table}[htbp]
    \centering
    \begin{minipage}{0.45\textwidth}
        \centering
        \caption{The quantitative results of semantic segmentation on Hypersim~\citep{roberts2021hypersim} testing set. Mean values are reported from 10 independent runs.}
        \label{tab:seg}
        \begin{tabular}{l|cc}
            \toprule
            Method & mIoU $\uparrow$ & mAcc $\uparrow$ \\ 
            \midrule
            Direct Adaption & 14.1 & 61.3\\ 
            \textbf{Lotus-G} & \textbf{21.2} & \textbf{65.6}\\ 
            \bottomrule
        \end{tabular}
        
    \end{minipage}
    \hfill
    \begin{minipage}{0.45\textwidth}
        \centering
        \caption{The quantitative results of diffuse reflectance prediction on Hypersim~\citep{roberts2021hypersim} testing set. Mean values are reported from 10 independent runs.}
        \label{tab:refl}
        \begin{tabular}{l|cc}
            \toprule
            Method & L1 $\downarrow$ & L2 $\downarrow$\\ 
            \midrule
            Direct Adaption & 0.198 & 0.206\\ 
            \textbf{Lotus-G} & \textbf{0.109} &\textbf{ 0.135} \\ 
            \bottomrule
        \end{tabular}
    \end{minipage}
    
\end{table}

To validate the generalization ability of our method on other dense prediction tasks, we further train it on semantic segmentation and diffuse reflectance prediction. Both tasks are trained using the training set of the Hypersim dataset~\citep{roberts2021hypersim} and evaluated on their corresponding test sets. For semantic segmentation, we report the mean intersection over union (mIoU) and mean accuracy (mAcc). For diffuse reflectance prediction, we evaluate using the L1 and L2 distances to the ground truth.
To enable fast evaluation, we randomly select 500 paired testing samples. In our experiments, we do not redesign any specific modules or loss functions for these tasks and maintain the original training protocol of Lotus unchanged.
As shown in Tab.~\ref{tab:seg} and Tab.~\ref{tab:refl}, we compare our method with the baseline, Direct Adaption (Fig. 4 in the main paper), to assess its effectiveness. The results show that our method outperforms the baseline across all metrics. Additionally, we provide qualitative visualizations for these two tasks in Fig.~\ref{fig:other_tasks}, demonstrating accurate and high-quality results. 
Both the quantitative and qualitative results indicate that our method, even without task-specific designs, can be effectively applied not only to geometric dense prediction tasks, as shown in the main paper, but also to semantic dense prediction tasks. 

\section{Frequency Domain Analysis of the Detail Preserver\\$\quad\ \ \ $Take Monocular Depth Estimation as An Example}
We use fast Fourier transform (FFT) to compute the Discrete Fourier Transform (DFT) of the input images and depth map estimations with and without Detail Preserver. The entire 2D frequency domains are divided into 8 frequency groups exponentially using the base of 2, \textit{i.e}., the first group covers the 2D frequency map in a circle with a radius of 2, the second group covers the annular region with radii from 2 to 4, the third group covers radii from 4 to 8, and so on. This exponential grouping allows us to analyze the frequency components across progressively larger ranges, capturing both low-frequency and high-frequency characteristics.

\begin{figure}[!ht]
\centering
    \begin{subfigure}[t]{0.48\linewidth}
    \includegraphics[width=\linewidth]{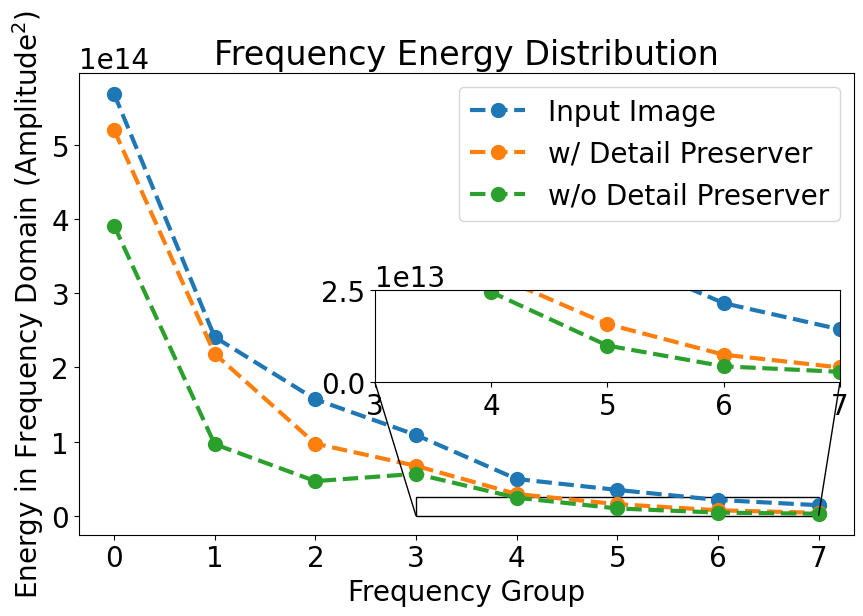}
    \caption{Frequency domain energy distribution comparisons among input image, and depth estimations \textit{w/} and \textit{w/o} Detail Preserver.}
    \label{fig:fft_1}
    \end{subfigure}
    \hfill
    \begin{subfigure}[t]{0.5\linewidth}
    \includegraphics[width = 1.0\linewidth]{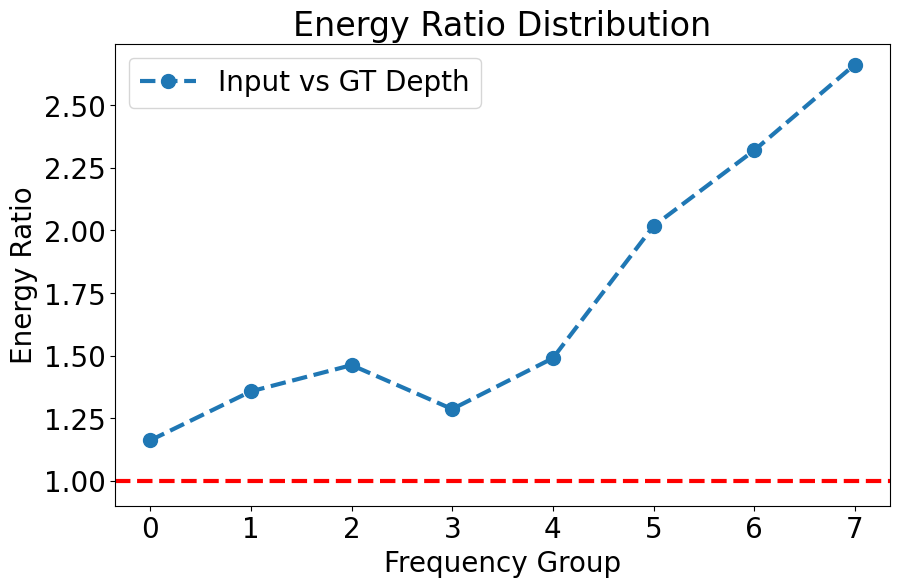}
    \caption{Frequency energy ratio \textbf{between input image and GT depth.}}
    \label{fig:fft_3}
    \end{subfigure}
    \newline
    \begin{subfigure}[t]{0.49\linewidth}
    \includegraphics[width=\linewidth]{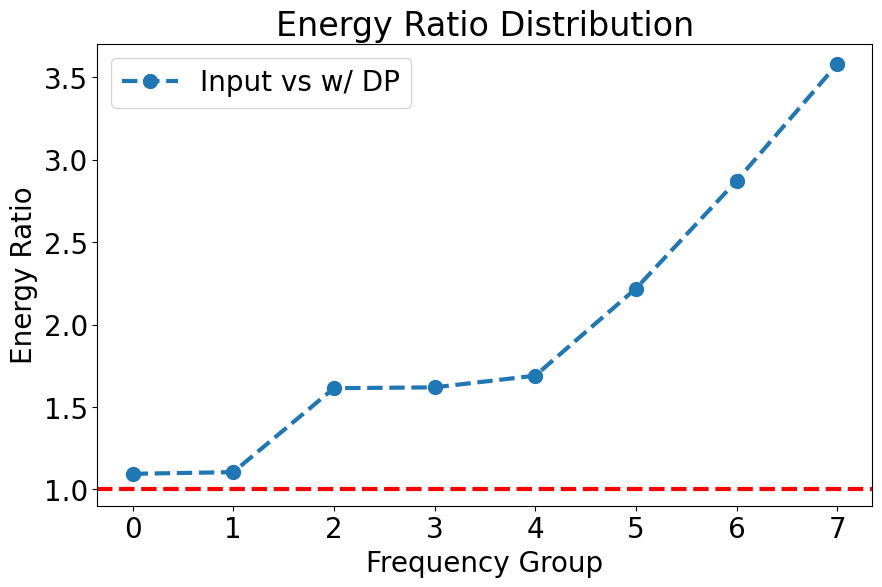}
    \caption{Frequency energy ratio \textbf{between input image and depth estimations \textit{w/} Detail Preserver.}}
    \label{fig:fft_4}
    \end{subfigure}
    \hfill
    \begin{subfigure}[t]{0.49\linewidth}
    \includegraphics[width=\linewidth]{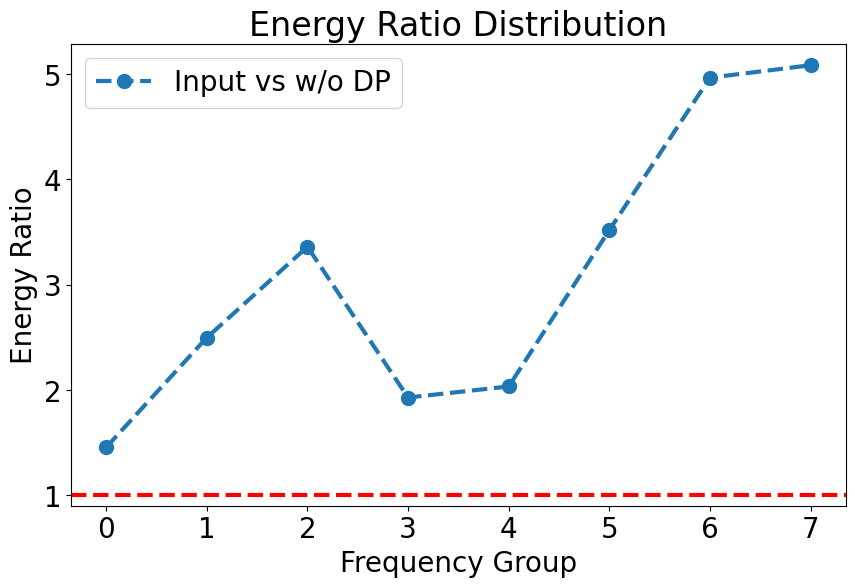}
    \caption{Frequency energy ratio \textbf{between the input image and depth estimations \textit{w/o} Detail Preserver.}}
    \label{fig:fft_5}
    \end{subfigure}
    \caption{\textbf{Frequency Domain Analysis of the Detail Preserver} We use Hypersim~\citep{roberts2021hypersim} dataset to transfer the input image and depth estimation \textit{w/} and \textit{w/o} Detail Preserver into 2D frequency domains, using FFT. 100 pairs of \{input image, depth estimation \textit{w/} Detail Preserver, depth estimation \textit{w/o} Detail Preserver\} are randomly selected for this frequency domain analysis. Hypersim is a photorealistic synthetic dataset. Not only can Hypersim offer dense GT labels without \texttt{None} areas (which is important during FFT), its depth annotations are much fine-grained compared with real-world datasets like NYUv2~\citep{silberman2012indoor} and KITTI~\cite{geiger2013vision}.}
    \label{fig:fft}
\end{figure}

In order to more clearly demonstrate the effect of our proposed Detail Preserver, we first analysis the experiments using Hypersim~\citep{roberts2021hypersim} dataset to display the difference in frequency domain energy between the details from both geometry and texture (the input images); and the details from purely the geometry (the GT depth maps).
As shown in Fig.~\ref{fig:fft_3}, the frequency domain energy between the input images and the depth annotations are plotted. Clearly we can see that the input images has much higher frequency energy in high-frequency areas, \textit{i.e.}, group 4, 5, 6, and 7, indicating that the details in surface textures mainly contribute to high-frequency energy; while the details in geometries, which can be expressed by depth maps, are mainly concentrated into (relative) middle and low frequency areas, \textit{i.e.}, group 0, 1, 2, and~3.

As shown in Fig.~\ref{fig:fft_1}
, collaborating with the Detail Preserver effectively drag the frequency domain energy of depth estimation to the input image, especially on middle and low frequency domains, \textit{i.e.}, the frequency group 0, 1, 2 and 3, highlighting the Detail Preserver’s effectiveness in enhancing the geometrical details that should be reflected into depth predictions, like the fences around roads and houses (Fig. 8 of our main paper).
While for high-frequency components, \textit{i.e.}, the frequency group 4, 5, 6, and 7, which may be primarily caused by the highly detailed textures, like the signs on the road and patterns on house surfaces, the energy in these areas between depth estimations with and without Detail Preserver is quite similar, indicating that the Detail Preserver does not copy this high-frequency and geometry-independent texture.

By comparing Fig.~\ref{fig:fft_3},~\ref{fig:fft_4} and~\ref{fig:fft_5} together, we can see that Detail Preserver effectively enhances the details of geometries. This insight is evident by this phenomenon: the frequency domain energy ratio between input and depth estimation \textit{w/} Detail Preserver, is closer to the frequency domain energy ratio between input and GT depth, compared with the frequency domain energy ratio between input and depth estimation \textit{w/o} Detail Preserver.

\section{The effect of different time-steps $t$ in one-step diffusion} 
In Sec. 4.2 of our main paper, we reduce the number of training time-steps of diffusion formulation to only one, and fixing the only time-step $t$ to $T$ following the diffusion formulation. In this section, we evaluate the effect of different time-steps $t$ in one-step diffusion, rather than exclusively fixing $t=T$, to validate that the rule of basic diffusion formulations should better be followed. Violating it will lead to performance degradation. 
As shown in Tab. \ref{tab:diff_t}, we conduct experiments on Hypersim dataset~\citep{roberts2021hypersim} and evaluated on NYUv2 dataset~\citep{silberman2012indoor}, without employing the detail preserver or mixture dataset training. The results indicate that the model performs best when $t=T$ ($t=1000$). Changing $t$ leads to a slight degradation in performance. 

\begin{table}[!ht]
\setlength{\tabcolsep}{14.pt}
    \centering
    \scriptsize
    \caption{\textbf{The effect of different time-steps $t$ in one-step diffusion.} In this experiment, the models are trained on Hypersim dataset~\citep{roberts2021hypersim} and evaluated on NYUv2 dataset~\citep{silberman2012indoor}, without employing the detail preserver or mixture dataset training. }
    \begin{tabular}{l|ccccc}
    \toprule
        Time-step & $t=1000$ & $t=750$ & $t=500$  & $t=250$  & $t=1$ \\ \midrule
        AbsRel $\downarrow$ & \textbf{5.587} & 5.631 & 5.727 & 5.663 & 5.737 \\ 
        $\delta 1$ $\uparrow$ & \textbf{96.272} & 96.165 & 96.087 & 96.141 & 96.080 \\ 
        \bottomrule
    \end{tabular}
    \label{tab:diff_t}
\end{table}

\begin{figure}[t]
    \vspace{17mm}
    \centering
    \includegraphics[width = \linewidth]{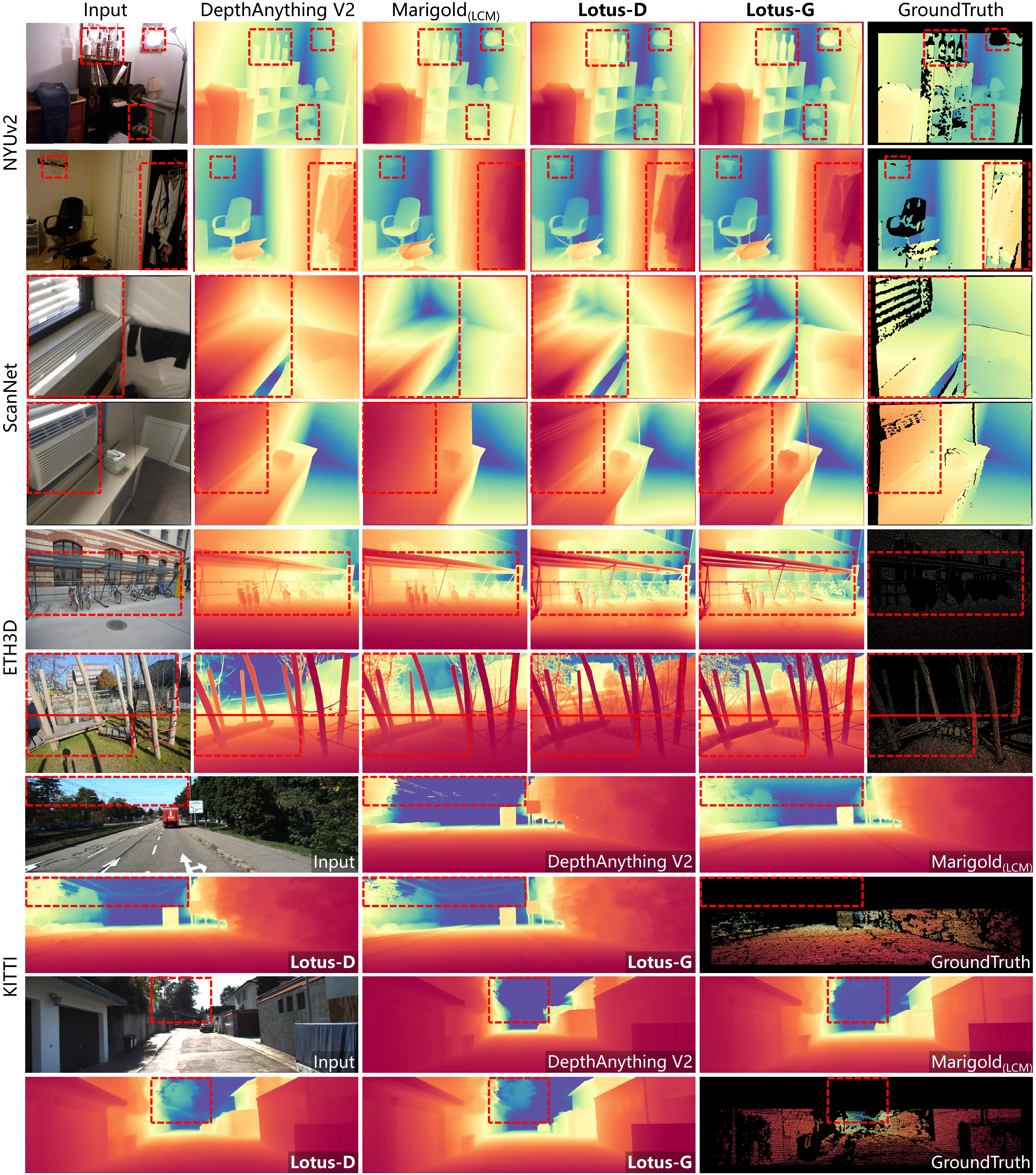}
    \caption{\textbf{Qualitative comparison on zero-shot affine-invariant depth estimation.} 
    Lotus demonstrates higher accuracy especially in detailed areas.
    }
    \vspace{-3mm}
    \label{fig:depth}
\end{figure}

\begin{figure}[t]
    \vspace{14mm}
    \centering
    \includegraphics[width = \linewidth]{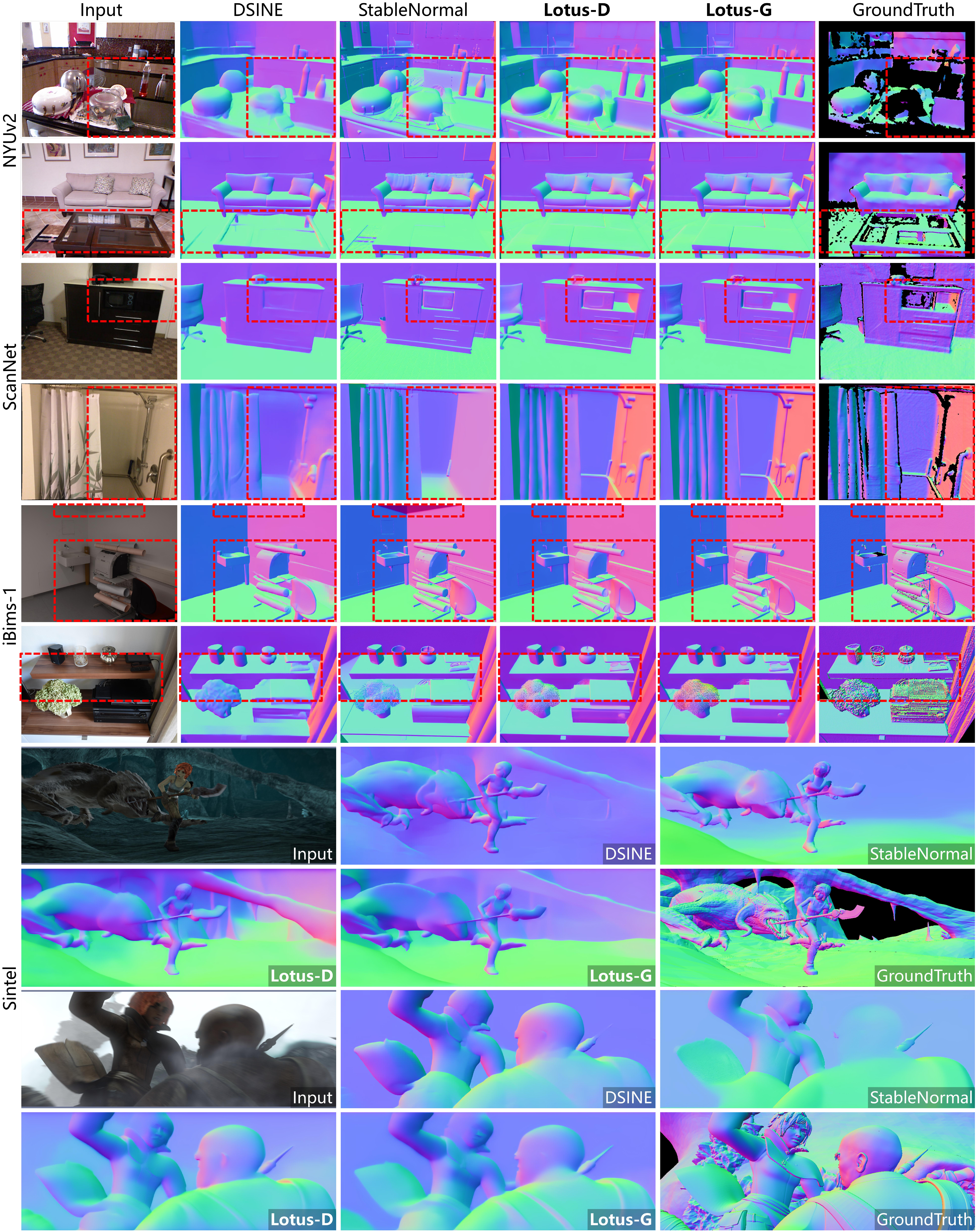}
    \caption{\textbf{Qualitative comparison on zero-shot surface normal estimation.} 
    Lotus offers improved accuracy particularly in complex regions.
    }
    \vspace{-3mm}
    \label{fig:normal}
\end{figure}

\section{Qualitative Comparisons}
\label{suppl:qualitative}
In Fig.~\ref{fig:depth}, we further compare the performance of our Lotus with other methods in detailed areas. The quantitative results obviously demonstrate that our method can produce much finer and more accurate depth predictions, particularly in complex regions with intricate structures, which sometimes cannot be reflected by the metrics. Also, as illustrated in Fig.~\ref{fig:normal}, Lotus consistently provides accurate surface normal predictions, effectively handling complex geometries and diverse environments, highlighting its robustness on fine-grained prediction.

\begin{figure}[t]
    \centering
    \includegraphics[width = \linewidth]{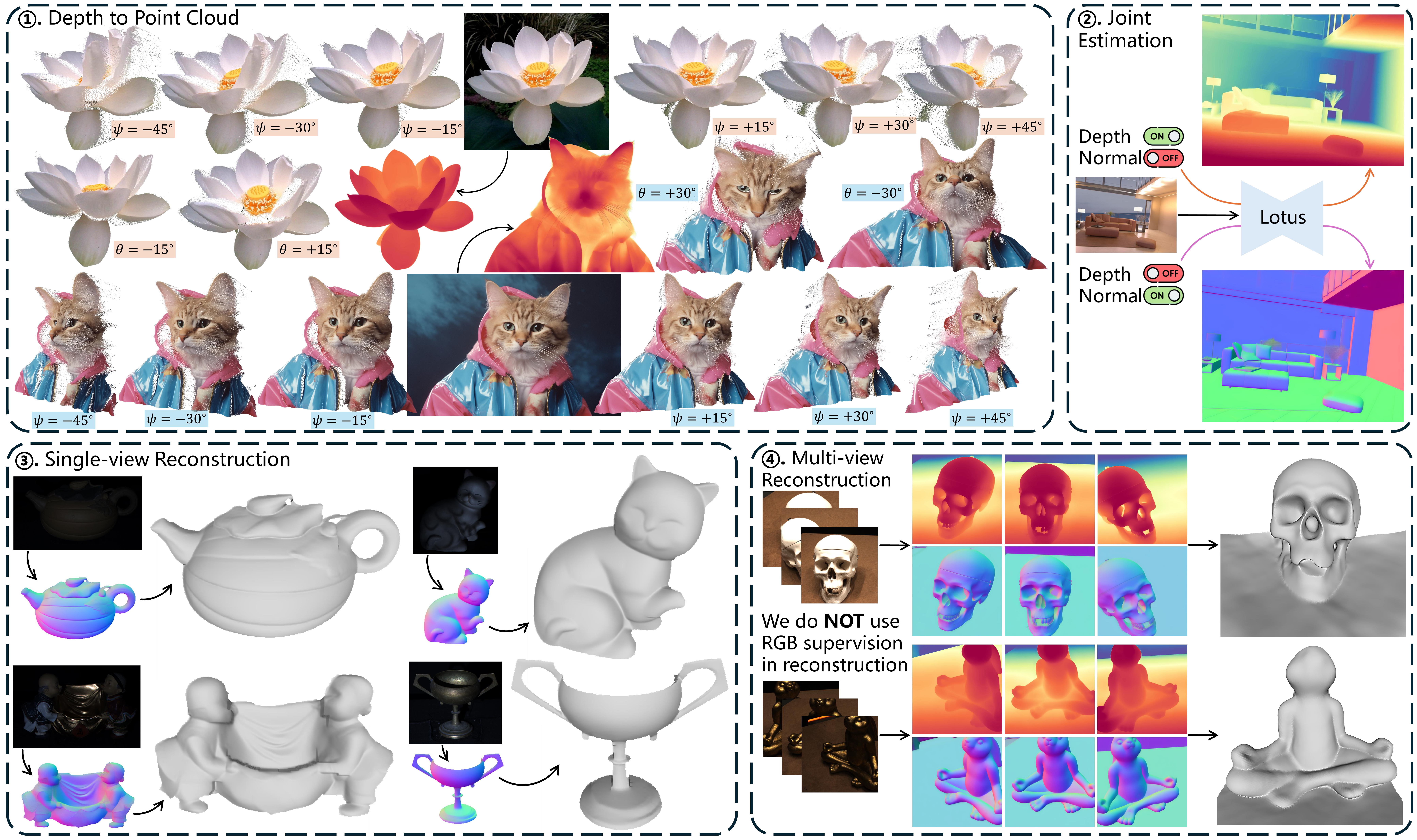}
    \caption{\textbf{Applications of Lotus.} 
    \ding{172} \textit{Depth to 3D Point Clouds}. \ding{173} \textit{Joint Estimation:} Simultaneous depth and normal estimation with $100\%$ shared parameters. \ding{174} \textit{Single-View Reconstruction:} Reconstructing 3D meshes from normal predictions. \ding{175} \textit{Multi-View Reconstruction:} Reconstructing high-quality meshes using depth/normal predictions \textbf{without RGB supervision}.}
    \label{fig:application}
\end{figure}

\section{Applications of Lotus}
\label{suppl:appl}
Thanks to its superiority, Lotus can seamlessly support a variety of applications. 
Fig.~\ref{fig:application} illustrates four key applications:
\ding{172} \textit{Depth to Point Cloud.} The depth maps estimated by Lotus are projected into 3D point clouds;
\ding{173} \textit{Joint Estimation.} By incorporating a task switcher, Lotus can perform multiple tasks simultaneously, such as joint depth and normal map estimation with $100\%$ shared network parameters; 
\ding{174} \textit{Single-View Reconstruction.} Using Lotus's normal predictions, high-quality meshes can be reconstructed through through Bilateral Normal Integration~\citep{bini2022cao}; 
\ding{175} \textit{Multi-View Reconstruction.} Leveraging per-view depth and normal predictions from Lotus, high-quality meshes can be reconstructed with MonoSDF~\citep{Yu2022MonoSDF}, \textbf{without RGB supervision}, showcasing Lotus's robustness and accurate spatial understanding. 
These applications emphasize the importance of Lotus in the field of computer vision. Its accuracy and efficiency will help in addressing increasingly complex problems.

\section{Future Work}
\label{suppl:future}
While we have applied Lotus to two geometric dense prediction tasks, it can be seamlessly adapted to other dense prediction tasks requiring per-pixel alignment with great potential, such as panoramic segmentation and image matting. Additionally, our performance is slightly behind DepthAnything~\citep{yang2024depth1} which utilizes large-scale training data. 
In the future, scaling up the training data, as reveal in Fig.~\ref{fig:T} and Tab.~\ref{tab:ablation} (``Mixture Dataset") of the main paper, has great potential to further enhance Lotus's performance.

\phantom{Some long invisible content to fill the page.}
\newpage
\phantom{Another long invisible content to fill the page.}
\newpage
\bibliography{iclr2024_conference}

\begin{thebibliography}{67}
\providecommand{\natexlab}[1]{#1}
\providecommand{\url}[1]{\texttt{#1}}
\expandafter\ifx\csname urlstyle\endcsname\relax
  \providecommand{\doi}[1]{doi: #1}\else
  \providecommand{\doi}{doi: \begingroup \urlstyle{rm}\Url}\fi

\bibitem[Bae \& Davison(2021)Bae and Davison]{Bae2021}
Gilwon Bae and Andrew~J Davison.
\newblock Aleatoric uncertainty in monocular surface normal estimation.
\newblock \emph{IEEE Transactions on Pattern Analysis and Machine Intelligence (TPAMI)}, pp.\  1472--1485, 2021.

\bibitem[Bae \& Davison(2024)Bae and Davison]{bae2024dsine}
Gilwon Bae and Andrew~J Davison.
\newblock Rethinking inductive biases for surface normal estimation.
\newblock \emph{IEEE Conference on Computer Vision and Pattern Recognition (CVPR)}, 2024.

\bibitem[Benny \& Wolf(2022)Benny and Wolf]{benny2022dynamic}
Yaniv Benny and Lior Wolf.
\newblock Dynamic dual-output diffusion models.
\newblock In \emph{Proceedings of the IEEE/CVF Conference on Computer Vision and Pattern Recognition}, pp.\  11482--11491, 2022.

\bibitem[Butler et~al.(2012)Butler, Wulff, Stanley, and Black]{butler2012naturalistic}
Daniel~J Butler, Jonas Wulff, Garrett~B Stanley, and Michael~J Black.
\newblock A naturalistic open source movie for optical flow evaluation.
\newblock In \emph{Computer Vision--ECCV 2012: 12th European Conference on Computer Vision, Florence, Italy, October 7-13, 2012, Proceedings, Part VI 12}, pp.\  611--625. Springer, 2012.

\bibitem[Cabon et~al.(2020)Cabon, Murray, and Humenberger]{cabon2020virtual}
Yohann Cabon, Naila Murray, and Martin Humenberger.
\newblock Virtual kitti 2.
\newblock \emph{arXiv preprint arXiv:2001.10773}, 2020.

\bibitem[Cao et~al.(2022)Cao, Santo, Shi, Okura, and Matsushita]{bini2022cao}
Xu~Cao, Hiroaki Santo, Boxin Shi, Fumio Okura, and Yasuyuki Matsushita.
\newblock Bilateral normal integration.
\newblock In \emph{ECCV}, 2022.

\bibitem[Chen et~al.(2023)Chen, Yu, Ge, Yao, Xie, Wu, Wang, Kwok, Luo, Lu, et~al.]{pixart}
Junsong Chen, Jincheng Yu, Chongjian Ge, Lewei Yao, Enze Xie, Yue Wu, Zhongdao Wang, James Kwok, Ping Luo, Huchuan Lu, et~al.
\newblock Pixart-$\alpha$: Fast training of diffusion transformer for photorealistic text-to-image synthesis.
\newblock \emph{arXiv preprint arXiv:2310.00426}, 2023.

\bibitem[Chen et~al.(2020)Chen, Qian, Fan, Kojima, Hamilton, and Deng]{chen2020oasis}
Weifeng Chen, Shengyi Qian, David Fan, Noriyuki Kojima, Max Hamilton, and Jia Deng.
\newblock Oasis: A large-scale dataset for single image 3d in the wild.
\newblock In \emph{Proceedings of the IEEE/CVF Conference on Computer Vision and Pattern Recognition}, pp.\  679--688, 2020.

\bibitem[Dai et~al.(2017)Dai, Chang, Savva, Halber, Funkhouser, and Nie{\ss}ner]{dai2017scannet}
Angela Dai, Angel~X Chang, Manolis Savva, Maciej Halber, Thomas Funkhouser, and Matthias Nie{\ss}ner.
\newblock Scannet: Richly-annotated 3d reconstructions of indoor scenes.
\newblock In \emph{Proceedings of the IEEE conference on computer vision and pattern recognition}, pp.\  5828--5839, 2017.

\bibitem[Du et~al.(2024)Du, Cheng, Luo, Qiu, Huang, Cheung, Cheng, and Fu]{du2024unlocking}
Wenyu Du, Shuang Cheng, Tongxu Luo, Zihan Qiu, Zeyu Huang, Ka~Chun Cheung, Reynold Cheng, and Jie Fu.
\newblock Unlocking continual learning abilities in language models.
\newblock \emph{arXiv preprint arXiv:2406.17245}, 2024.

\bibitem[Eftekhar et~al.(2021)Eftekhar, Sax, Malik, and Zamir]{eftekhar2021omnidata}
Ainaz Eftekhar, Alexander Sax, Jitendra Malik, and Amir Zamir.
\newblock Omnidata: A scalable pipeline for making multi-task mid-level vision datasets from 3d scans.
\newblock In \emph{Proceedings of the IEEE/CVF International Conference on Computer Vision}, pp.\  10786--10796, 2021.

\bibitem[Eigen et~al.(2014)Eigen, Puhrsch, and Fergus]{eigen2014depth}
David Eigen, Christian Puhrsch, and Rob Fergus.
\newblock Depth map prediction from a single image using a multi-scale deep network.
\newblock \emph{Advances in neural information processing systems}, 27, 2014.

\bibitem[Fu et~al.(2018)Fu, Gong, Wang, Batmanghelich, and Tao]{fu2018deep}
Huan Fu, Mingming Gong, Chaohui Wang, Kayhan Batmanghelich, and Dacheng Tao.
\newblock Deep ordinal regression network for monocular depth estimation.
\newblock In \emph{Proceedings of the IEEE conference on computer vision and pattern recognition}, pp.\  2002--2011, 2018.

\bibitem[Fu et~al.(2024)Fu, Yin, Hu, Wang, Ma, Tan, Shen, Lin, and Long]{fu2024geowizard}
Xiao Fu, Wei Yin, Mu~Hu, Kaixuan Wang, Yuexin Ma, Ping Tan, Shaojie Shen, Dahua Lin, and Xiaoxiao Long.
\newblock Geowizard: Unleashing the diffusion priors for 3d geometry estimation from a single image.
\newblock \emph{arXiv preprint arXiv:2403.12013}, 2024.

\bibitem[Garcia et~al.(2024)Garcia, Zeid, Schmidt, de~Geus, Hermans, and Leibe]{garcia2024fine}
Gonzalo~Martin Garcia, Karim~Abou Zeid, Christian Schmidt, Daan de~Geus, Alexander Hermans, and Bastian Leibe.
\newblock Fine-tuning image-conditional diffusion models is easier than you think.
\newblock \emph{arXiv preprint arXiv:2409.11355}, 2024.

\bibitem[Geiger et~al.(2013)Geiger, Lenz, Stiller, and Urtasun]{geiger2013vision}
Andreas Geiger, Philip Lenz, Christoph Stiller, and Raquel Urtasun.
\newblock Vision meets robotics: The kitti dataset.
\newblock \emph{The International Journal of Robotics Research}, 32\penalty0 (11):\penalty0 1231--1237, 2013.

\bibitem[Goodfellow et~al.(2014)Goodfellow, Pouget-Abadie, Mirza, Xu, Warde-Farley, Ozair, Courville, and Bengio]{goodfellow2014generative}
Ian Goodfellow, Jean Pouget-Abadie, Mehdi Mirza, Bing Xu, David Warde-Farley, Sherjil Ozair, Aaron Courville, and Yoshua Bengio.
\newblock Generative adversarial nets.
\newblock \emph{Advances in neural information processing systems}, 27, 2014.

\bibitem[He et~al.(2022)He, Zhou, Zhang, Peng, Shen, Sun, Chen, and Ji]{pixelfolder}
Jing He, Yiyi Zhou, Qi~Zhang, Jun Peng, Yunhang Shen, Xiaoshuai Sun, Chao Chen, and Rongrong Ji.
\newblock Pixelfolder: An efficient progressive pixel synthesis network for image generation.
\newblock \emph{arXiv preprint arXiv:2204.00833}, 2022.

\bibitem[He et~al.(2024)He, Li, Hu, Shen, Cai, Qiu, and Chen]{he2024disenvisioner}
Jing He, Haodong Li, Yongzhe Hu, Guibao Shen, Yingjie Cai, Weichao Qiu, and Ying-Cong Chen.
\newblock Disenvisioner: Disentangled and enriched visual prompt for customized image generation.
\newblock \emph{arXiv preprint arXiv:2410.02067}, 2024.

\bibitem[Ho et~al.(2020)Ho, Jain, and Abbeel]{ho2020denoising}
Jonathan Ho, Ajay Jain, and Pieter Abbeel.
\newblock Denoising diffusion probabilistic models.
\newblock \emph{Advances in neural information processing systems}, 33:\penalty0 6840--6851, 2020.

\bibitem[Hu et~al.(2024)Hu, Yin, Zhang, Cai, Long, Chen, Wang, Yu, Shen, and Shen]{hu2024metric3d}
Mu~Hu, Wei Yin, Chi Zhang, Zhipeng Cai, Xiaoxiao Long, Hao Chen, Kaixuan Wang, Gang Yu, Chunhua Shen, and Shaojie Shen.
\newblock Metric3d v2: A versatile monocular geometric foundation model for zero-shot metric depth and surface normal estimation.
\newblock \emph{arXiv preprint arXiv:2404.15506}, 2024.

\bibitem[Hu et~al.(2023)Hu, Yang, Chen, Li, Sima, Zhu, Chai, Du, Lin, Wang, et~al.]{hu2023planning}
Yihan Hu, Jiazhi Yang, Li~Chen, Keyu Li, Chonghao Sima, Xizhou Zhu, Siqi Chai, Senyao Du, Tianwei Lin, Wenhai Wang, et~al.
\newblock Planning-oriented autonomous driving.
\newblock In \emph{Proceedings of the IEEE/CVF Conference on Computer Vision and Pattern Recognition}, pp.\  17853--17862, 2023.

\bibitem[Huang et~al.(2024)Huang, Yu, Chen, Geiger, and Gao]{Huang2DGS2024}
Binbin Huang, Zehao Yu, Anpei Chen, Andreas Geiger, and Shenghua Gao.
\newblock 2d gaussian splatting for geometrically accurate radiance fields.
\newblock In \emph{SIGGRAPH 2024 Conference Papers}. Association for Computing Machinery, 2024.
\newblock \doi{10.1145/3641519.3657428}.

\bibitem[Kar et~al.(2022)Kar, Yeo, Atanov, and Zamir]{kar20223d}
O{\u{g}}uzhan~Fatih Kar, Teresa Yeo, Andrei Atanov, and Amir Zamir.
\newblock 3d common corruptions and data augmentation.
\newblock In \emph{Proceedings of the IEEE/CVF Conference on Computer Vision and Pattern Recognition}, pp.\  18963--18974, 2022.

\bibitem[Karras et~al.(2019)Karras, Laine, and Aila]{StyleGAN1}
Tero Karras, Samuli Laine, and Timo Aila.
\newblock A style-based generator architecture for generative adversarial networks.
\newblock In \emph{Proceedings of the IEEE/CVF conference on computer vision and pattern recognition}, pp.\  4401--4410, 2019.

\bibitem[Karras et~al.(2020)Karras, Laine, Aittala, Hellsten, Lehtinen, and Aila]{StyleGAN2}
Tero Karras, Samuli Laine, Miika Aittala, Janne Hellsten, Jaakko Lehtinen, and Timo Aila.
\newblock Analyzing and improving the image quality of stylegan.
\newblock In \emph{Proceedings of the IEEE/CVF conference on computer vision and pattern recognition}, pp.\  8110--8119, 2020.

\bibitem[Karras et~al.(2021)Karras, Aittala, Laine, H{\"a}rk{\"o}nen, Hellsten, Lehtinen, and Aila]{StyleGAN3}
Tero Karras, Miika Aittala, Samuli Laine, Erik H{\"a}rk{\"o}nen, Janne Hellsten, Jaakko Lehtinen, and Timo Aila.
\newblock Alias-free generative adversarial networks.
\newblock \emph{Advances in Neural Information Processing Systems}, 34:\penalty0 852--863, 2021.

\bibitem[Ke et~al.(2024)Ke, Obukhov, Huang, Metzger, Daudt, and Schindler]{ke2024repurposing}
Bingxin Ke, Anton Obukhov, Shengyu Huang, Nando Metzger, Rodrigo~Caye Daudt, and Konrad Schindler.
\newblock Repurposing diffusion-based image generators for monocular depth estimation.
\newblock In \emph{Proceedings of the IEEE/CVF Conference on Computer Vision and Pattern Recognition}, pp.\  9492--9502, 2024.

\bibitem[Koch et~al.(2018)Koch, Liebel, Fraundorfer, and Korner]{koch2018evaluation}
Tobias Koch, Lukas Liebel, Friedrich Fraundorfer, and Marco Korner.
\newblock Evaluation of cnn-based single-image depth estimation methods.
\newblock In \emph{Proceedings of the European Conference on Computer Vision (ECCV) Workshops}, pp.\  0--0, 2018.

\bibitem[Lee et~al.(2024)Lee, Tseng, and Yang]{lee2024exploiting}
Hsin-Ying Lee, Hung-Yu Tseng, and Ming-Hsuan Yang.
\newblock Exploiting diffusion prior for generalizable dense prediction.
\newblock In \emph{Proceedings of the IEEE/CVF Conference on Computer Vision and Pattern Recognition}, pp.\  7861--7871, 2024.

\bibitem[Lee et~al.(2019)Lee, Han, Ko, and Suh]{lee2019big}
Jin~Han Lee, Myung-Kyu Han, Dong~Wook Ko, and Il~Hong Suh.
\newblock From big to small: Multi-scale local planar guidance for monocular depth estimation.
\newblock \emph{arXiv preprint arXiv:1907.10326}, 2019.

\bibitem[Lei et~al.(2024)Lei, Weng, Harley, Guibas, and Daniilidis]{lei2024mosca}
Jiahui Lei, Yijia Weng, Adam Harley, Leonidas Guibas, and Kostas Daniilidis.
\newblock Mosca: Dynamic gaussian fusion from casual videos via 4d motion scaffolds.
\newblock \emph{arXiv preprint arXiv:2405.17421}, 2024.

\bibitem[Long et~al.(2024)Long, Guo, Lin, Liu, Dou, Liu, Ma, Zhang, Habermann, Theobalt, et~al.]{long2024wonder3d}
Xiaoxiao Long, Yuan-Chen Guo, Cheng Lin, Yuan Liu, Zhiyang Dou, Lingjie Liu, Yuexin Ma, Song-Hai Zhang, Marc Habermann, Christian Theobalt, et~al.
\newblock Wonder3d: Single image to 3d using cross-domain diffusion.
\newblock In \emph{Proceedings of the IEEE/CVF Conference on Computer Vision and Pattern Recognition}, pp.\  9970--9980, 2024.

\bibitem[Nichol et~al.(2021)Nichol, Dhariwal, Ramesh, Shyam, Mishkin, McGrew, Sutskever, and Chen]{nichol2021glide}
Alex Nichol, Prafulla Dhariwal, Aditya Ramesh, Pranav Shyam, Pamela Mishkin, Bob McGrew, Ilya Sutskever, and Mark Chen.
\newblock Glide: Towards photorealistic image generation and editing with text-guided diffusion models.
\newblock \emph{arXiv preprint arXiv:2112.10741}, 2021.

\bibitem[Ramesh et~al.(2021)Ramesh, Pavlov, Goh, Gray, Voss, Radford, Chen, and Sutskever]{dalle}
Aditya Ramesh, Mikhail Pavlov, Gabriel Goh, Scott Gray, Chelsea Voss, Alec Radford, Mark Chen, and Ilya Sutskever.
\newblock Zero-shot text-to-image generation.
\newblock In \emph{International Conference on Machine Learning}, pp.\  8821--8831. PMLR, 2021.

\bibitem[Ramesh et~al.(2022)Ramesh, Dhariwal, Nichol, Chu, and Chen]{unclip}
Aditya Ramesh, Prafulla Dhariwal, Alex Nichol, Casey Chu, and Mark Chen.
\newblock Hierarchical text-conditional image generation with clip latents.
\newblock \emph{arXiv preprint arXiv:2204.06125}, 1\penalty0 (2):\penalty0 3, 2022.

\bibitem[Ranftl et~al.(2020)Ranftl, Lasinger, Hafner, Schindler, and Koltun]{ranftl2020towards}
Ren{\'e} Ranftl, Katrin Lasinger, David Hafner, Konrad Schindler, and Vladlen Koltun.
\newblock Towards robust monocular depth estimation: Mixing datasets for zero-shot cross-dataset transfer.
\newblock \emph{IEEE transactions on pattern analysis and machine intelligence}, 44\penalty0 (3):\penalty0 1623--1637, 2020.

\bibitem[Ranftl et~al.(2021)Ranftl, Bochkovskiy, and Koltun]{ranftl2021vision}
Ren{\'e} Ranftl, Alexey Bochkovskiy, and Vladlen Koltun.
\newblock Vision transformers for dense prediction.
\newblock In \emph{Proceedings of the IEEE/CVF international conference on computer vision}, pp.\  12179--12188, 2021.

\bibitem[Roberts et~al.(2021)Roberts, Ramapuram, Ranjan, Kumar, Bautista, Paczan, Webb, and Susskind]{roberts2021hypersim}
Mike Roberts, Jason Ramapuram, Anurag Ranjan, Atulit Kumar, Miguel~Angel Bautista, Nathan Paczan, Russ Webb, and Joshua~M Susskind.
\newblock Hypersim: A photorealistic synthetic dataset for holistic indoor scene understanding.
\newblock In \emph{Proceedings of the IEEE/CVF international conference on computer vision}, pp.\  10912--10922, 2021.

\bibitem[Rombach et~al.(2022)Rombach, Blattmann, Lorenz, Esser, and Ommer]{rombach2022high}
Robin Rombach, Andreas Blattmann, Dominik Lorenz, Patrick Esser, and Bj{\"o}rn Ommer.
\newblock High-resolution image synthesis with latent diffusion models.
\newblock In \emph{Proceedings of the IEEE/CVF conference on computer vision and pattern recognition}, pp.\  10684--10695, 2022.

\bibitem[Ronneberger et~al.(2015)Ronneberger, Fischer, and Brox]{ronneberger2015u}
Olaf Ronneberger, Philipp Fischer, and Thomas Brox.
\newblock U-net: Convolutional networks for biomedical image segmentation.
\newblock In \emph{Medical image computing and computer-assisted intervention--MICCAI 2015: 18th international conference, Munich, Germany, October 5-9, 2015, proceedings, part III 18}, pp.\  234--241. Springer, 2015.

\bibitem[Saharia et~al.(2022)Saharia, Chan, Saxena, Li, Whang, Denton, Ghasemipour, Gontijo~Lopes, Karagol~Ayan, Salimans, et~al.]{imagen}
Chitwan Saharia, William Chan, Saurabh Saxena, Lala Li, Jay Whang, Emily~L Denton, Kamyar Ghasemipour, Raphael Gontijo~Lopes, Burcu Karagol~Ayan, Tim Salimans, et~al.
\newblock Photorealistic text-to-image diffusion models with deep language understanding.
\newblock \emph{Advances in Neural Information Processing Systems}, 35:\penalty0 36479--36494, 2022.

\bibitem[Salimans \& Ho(2022)Salimans and Ho]{salimans2022progressive}
Tim Salimans and Jonathan Ho.
\newblock Progressive distillation for fast sampling of diffusion models.
\newblock \emph{arXiv preprint arXiv:2202.00512}, 2022.

\bibitem[Schops et~al.(2017)Schops, Schonberger, Galliani, Sattler, Schindler, Pollefeys, and Geiger]{schops2017multi}
Thomas Schops, Johannes~L Schonberger, Silvano Galliani, Torsten Sattler, Konrad Schindler, Marc Pollefeys, and Andreas Geiger.
\newblock A multi-view stereo benchmark with high-resolution images and multi-camera videos.
\newblock In \emph{Proceedings of the IEEE conference on computer vision and pattern recognition}, pp.\  3260--3269, 2017.

\bibitem[Schuhmann et~al.(2022)Schuhmann, Beaumont, Vencu, Gordon, Wightman, Cherti, Coombes, Katta, Mullis, Wortsman, et~al.]{schuhmann2022laion}
Christoph Schuhmann, Romain Beaumont, Richard Vencu, Cade Gordon, Ross Wightman, Mehdi Cherti, Theo Coombes, Aarush Katta, Clayton Mullis, Mitchell Wortsman, et~al.
\newblock Laion-5b: An open large-scale dataset for training next generation image-text models.
\newblock \emph{Advances in Neural Information Processing Systems}, 35:\penalty0 25278--25294, 2022.

\bibitem[Silberman et~al.(2012)Silberman, Hoiem, Kohli, and Fergus]{silberman2012indoor}
Nathan Silberman, Derek Hoiem, Pushmeet Kohli, and Rob Fergus.
\newblock Indoor segmentation and support inference from rgbd images.
\newblock In \emph{Computer Vision--ECCV 2012: 12th European Conference on Computer Vision, Florence, Italy, October 7-13, 2012, Proceedings, Part V 12}, pp.\  746--760. Springer, 2012.

\bibitem[Song et~al.(2020)Song, Meng, and Ermon]{song2020denoising}
Jiaming Song, Chenlin Meng, and Stefano Ermon.
\newblock Denoising diffusion implicit models.
\newblock \emph{arXiv preprint arXiv:2010.02502}, 2020.

\bibitem[Song et~al.(2024)Song, Lei, Wang, Liu, and Daniilidis]{song2024track}
Yunzhou Song, Jiahui Lei, Ziyun Wang, Lingjie Liu, and Kostas Daniilidis.
\newblock Track everything everywhere fast and robustly, 2024.

\bibitem[Vasiljevic et~al.(2019)Vasiljevic, Kolkin, Zhang, Luo, Wang, Dai, Daniele, Mostajabi, Basart, Walter, et~al.]{vasiljevic2019diode}
Igor Vasiljevic, Nick Kolkin, Shanyi Zhang, Ruotian Luo, Haochen Wang, Falcon~Z Dai, Andrea~F Daniele, Mohammadreza Mostajabi, Steven Basart, Matthew~R Walter, et~al.
\newblock Diode: A dense indoor and outdoor depth dataset.
\newblock \emph{arXiv preprint arXiv:1908.00463}, 2019.

\bibitem[Wang et~al.(2024)Wang, Ye, Gao, Austin, Li, and Kanazawa]{wang2024shape}
Qianqian Wang, Vickie Ye, Hang Gao, Jake Austin, Zhengqi Li, and Angjoo Kanazawa.
\newblock Shape of motion: 4d reconstruction from a single video.
\newblock \emph{arXiv preprint arXiv:2407.13764}, 2024.

\bibitem[Xiao et~al.(2024)Xiao, Wang, Zhang, Xue, Peng, Shen, and Zhou]{SpatialTracker}
Yuxi Xiao, Qianqian Wang, Shangzhan Zhang, Nan Xue, Sida Peng, Yujun Shen, and Xiaowei Zhou.
\newblock Spatialtracker: Tracking any 2d pixels in 3d space.
\newblock In \emph{Proceedings of the IEEE/CVF Conference on Computer Vision and Pattern Recognition (CVPR)}, 2024.

\bibitem[Xu et~al.(2024)Xu, Ge, Liu, Fan, Xie, Zhao, Chen, and Shen]{xu2024diffusion}
Guangkai Xu, Yongtao Ge, Mingyu Liu, Chengxiang Fan, Kangyang Xie, Zhiyue Zhao, Hao Chen, and Chunhua Shen.
\newblock Diffusion models trained with large data are transferable visual models.
\newblock \emph{arXiv preprint arXiv:2403.06090}, 2024.

\bibitem[Xu et~al.(2018)Xu, Zhang, Huang, Zhang, Gan, Huang, and He]{xu2018attngan}
Tao Xu, Pengchuan Zhang, Qiuyuan Huang, Han Zhang, Zhe Gan, Xiaolei Huang, and Xiaodong He.
\newblock Attngan: Fine-grained text to image generation with attentional generative adversarial networks.
\newblock In \emph{Proceedings of the IEEE conference on computer vision and pattern recognition}, pp.\  1316--1324, 2018.

\bibitem[Yang et~al.(2024{\natexlab{a}})Yang, Kang, Huang, Xu, Feng, and Zhao]{yang2024depth1}
Lihe Yang, Bingyi Kang, Zilong Huang, Xiaogang Xu, Jiashi Feng, and Hengshuang Zhao.
\newblock Depth anything: Unleashing the power of large-scale unlabeled data.
\newblock In \emph{Proceedings of the IEEE/CVF Conference on Computer Vision and Pattern Recognition}, pp.\  10371--10381, 2024{\natexlab{a}}.

\bibitem[Yang et~al.(2024{\natexlab{b}})Yang, Kang, Huang, Zhao, Xu, Feng, and Zhao]{yang2024depth2}
Lihe Yang, Bingyi Kang, Zilong Huang, Zhen Zhao, Xiaogang Xu, Jiashi Feng, and Hengshuang Zhao.
\newblock Depth anything v2.
\newblock \emph{arXiv preprint arXiv:2406.09414}, 2024{\natexlab{b}}.

\bibitem[Ye et~al.(2024)Ye, Qiu, Gu, Zuo, Wu, Dong, Bo, Xiu, and Han]{ye2024stablenormal}
Chongjie Ye, Lingteng Qiu, Xiaodong Gu, Qi~Zuo, Yushuang Wu, Zilong Dong, Liefeng Bo, Yuliang Xiu, and Xiaoguang Han.
\newblock Stablenormal: Reducing diffusion variance for stable and sharp normal.
\newblock \emph{arXiv preprint arXiv:2406.16864}, 2024.

\bibitem[Yin et~al.(2021{\natexlab{a}})Yin, Liu, and Shen]{yin2021virtual}
Wei Yin, Yifan Liu, and Chunhua Shen.
\newblock Virtual normal: Enforcing geometric constraints for accurate and robust depth prediction.
\newblock \emph{IEEE Transactions on Pattern Analysis and Machine Intelligence}, 44\penalty0 (10):\penalty0 7282--7295, 2021{\natexlab{a}}.

\bibitem[Yin et~al.(2021{\natexlab{b}})Yin, Zhang, Wang, Niklaus, Mai, Chen, and Shen]{yin2021learning}
Wei Yin, Jianming Zhang, Oliver Wang, Simon Niklaus, Long Mai, Simon Chen, and Chunhua Shen.
\newblock Learning to recover 3d scene shape from a single image.
\newblock In \emph{Proceedings of the IEEE/CVF Conference on Computer Vision and Pattern Recognition}, pp.\  204--213, 2021{\natexlab{b}}.

\bibitem[Yin et~al.(2023)Yin, Zhang, Chen, Cai, Yu, Wang, Chen, and Shen]{yin2023metric3d}
Wei Yin, Chi Zhang, Hao Chen, Zhipeng Cai, Gang Yu, Kaixuan Wang, Xiaozhi Chen, and Chunhua Shen.
\newblock Metric3d: Towards zero-shot metric 3d prediction from a single image.
\newblock In \emph{Proceedings of the IEEE/CVF International Conference on Computer Vision}, pp.\  9043--9053, 2023.

\bibitem[Yu et~al.(2022)Yu, Peng, Niemeyer, Sattler, and Geiger]{Yu2022MonoSDF}
Zehao Yu, Songyou Peng, Michael Niemeyer, Torsten Sattler, and Andreas Geiger.
\newblock Monosdf: Exploring monocular geometric cues for neural implicit surface reconstruction.
\newblock \emph{Advances in Neural Information Processing Systems (NeurIPS)}, 2022.

\bibitem[Yuan et~al.(2022)Yuan, Gu, Dai, Zhu, and Tan]{yuan2022neural}
Weihao Yuan, Xiaodong Gu, Zuozhuo Dai, Siyu Zhu, and Ping Tan.
\newblock Neural window fully-connected crfs for monocular depth estimation.
\newblock In \emph{Proceedings of the IEEE/CVF conference on computer vision and pattern recognition}, pp.\  3916--3925, 2022.

\bibitem[Yurtsever et~al.(2020)Yurtsever, Lambert, Carballo, and Takeda]{yurtsever2020survey}
Ekim Yurtsever, Jacob Lambert, Alexander Carballo, and Kazuya Takeda.
\newblock A survey of autonomous driving: Common practices and emerging technologies.
\newblock \emph{IEEE access}, 8:\penalty0 58443--58469, 2020.

\bibitem[Zhai et~al.(2023)Zhai, Tong, Li, Cai, Qu, Lee, and Ma]{zhai2023investigating}
Yuexiang Zhai, Shengbang Tong, Xiao Li, Mu~Cai, Qing Qu, Yong~Jae Lee, and Yi~Ma.
\newblock Investigating the catastrophic forgetting in multimodal large language models.
\newblock \emph{arXiv preprint arXiv:2309.10313}, 2023.

\bibitem[Zhang et~al.(2022)Zhang, Yin, Wang, Yu, Fu, and Shen]{zhang2022hierarchical}
Chi Zhang, Wei Yin, Billzb Wang, Gang Yu, Bin Fu, and Chunhua Shen.
\newblock Hierarchical normalization for robust monocular depth estimation.
\newblock \emph{Advances in Neural Information Processing Systems}, 35:\penalty0 14128--14139, 2022.

\bibitem[Zhang et~al.(2017)Zhang, Xu, Li, Zhang, Wang, Huang, and Metaxas]{zhang2017stackgan}
Han Zhang, Tao Xu, Hongsheng Li, Shaoting Zhang, Xiaogang Wang, Xiaolei Huang, and Dimitris~N Metaxas.
\newblock Stackgan: Text to photo-realistic image synthesis with stacked generative adversarial networks.
\newblock In \emph{Proceedings of the IEEE international conference on computer vision}, pp.\  5907--5915, 2017.

\bibitem[Zhang et~al.(2018)Zhang, Xu, Li, Zhang, Wang, Huang, and Metaxas]{zhang2018stackgan++}
Han Zhang, Tao Xu, Hongsheng Li, Shaoting Zhang, Xiaogang Wang, Xiaolei Huang, and Dimitris~N Metaxas.
\newblock Stackgan++: Realistic image synthesis with stacked generative adversarial networks.
\newblock \emph{IEEE transactions on pattern analysis and machine intelligence}, 41\penalty0 (8):\penalty0 1947--1962, 2018.

\bibitem[Zhang et~al.(2021)Zhang, Koh, Baldridge, Lee, and Yang]{zhang2021cross}
Han Zhang, Jing~Yu Koh, Jason Baldridge, Honglak Lee, and Yinfei Yang.
\newblock Cross-modal contrastive learning for text-to-image generation.
\newblock In \emph{Proceedings of the IEEE/CVF conference on computer vision and pattern recognition}, pp.\  833--842, 2021.

\end{thebibliography}
\bibliographystyle{iclr2024_conference}


\end{document}